\documentclass[manuscript,nonacm]{acmart} 

\AtBeginDocument{%
  }

\usepackage{booktabs}
\usepackage{multirow}
\usepackage{multicol}
\usepackage{xcolor}
\usepackage{amsmath}
\usepackage{wrapfig}
\usepackage{amsfonts}
\usepackage{amssymb}
\usepackage[linesnumbered, ruled]{algorithm2e}
\usepackage{amsthm}
\usepackage{booktabs}
\usepackage{colortbl}
\usepackage{multirow}
\usepackage{longtable}
\usepackage{enumitem}
\usepackage{graphicx}
\usepackage{subcaption}
\usepackage{todonotes}
\usepackage{hyperref}
\usepackage{color}
\usepackage{pifont}
\usepackage{lscape}
\usepackage{makecell}
\def\rot{\rotatebox}

\usepackage{csquotes}

\usepackage[nolist]{acronym}

\begin{document}

\title{Gender Bias in Explainability: Investigating Performance Disparity in Post-hoc Methods}

\author{Mahdi Dhaini}
\affiliation{%
  \institution{Technical University of Munich, School of Computation, Information and Technology, Department of Computer Science, Munich}
  \country{Germany}
}
\email{mahdi.dhaini@tum.de}
 
\author{Ege Erdogan}
\affiliation{%
    \institution{Technical University of Munich, { School of Computation, Information and Technology, Department of Computer Science,} Munich}
  \country{Germany}
}
\email{ege.erdogan@tum.de}

\author{Nils Feldhus}
\affiliation{%
   \institution{Technische Universit\"at Berlin, BIFOLD – Berlin Institute for the Foundations of Learning and Data, German Research Center for Artificial Intelligence (DFKI), Berlin}
   \country{Germany}
}
\email{nils.feldhus@dfki.de}

\author{Gjergji Kasneci}
\affiliation{%
  \institution{Technical University of Munich, { School of Computation, Information and Technology, Department of Computer Science,} Munich}
  \country{Germany}
}
\email{gjergji.kasneci@tum.de}


\renewcommand{\shortauthors}{Dhaini et al.}

\begin{abstract}

  While research on applications and evaluations of explanation methods continues to expand, fairness of the explanation methods concerning disparities in their performance across subgroups remains an often overlooked aspect. In this paper, we address this gap by showing that, across three tasks and five language models, widely used post-hoc feature attribution methods exhibit significant gender disparity with respect to their faithfulness, robustness, and complexity. These disparities persist even when the models are pre-trained or fine-tuned on particularly unbiased datasets, indicating that the disparities we observe are not merely consequences of biased training data. Our results highlight the importance of addressing disparities in explanations when developing and applying explainability methods, as these can lead to biased outcomes against certain subgroups, with particularly critical implications in high-stakes contexts. Furthermore, our findings underscore the importance of incorporating the fairness of explanations, alongside overall model fairness and explainability, as a requirement in regulatory frameworks.  

\end{abstract}



\maketitle

\section{Introduction} \label{sec:intro}
   
    Pre-trained language models (PLMs) are increasingly used in various natural language processing (NLP) tasks but are often hard-to-understand black boxes, which makes the problems of \textit{explaining PLMs} and \textit{evaluating those explanations} highly valuable. The growing demand to understand how PLMs generate their outputs has led to the increased adoption of Explainable AI methods in NLP. Explainable NLP, in particular, focuses on developing and applying techniques to interpret the inner workings and predictions of NLP models, including PLMs. Model-agnostic \textit{post-hoc} feature importance methods have been particularly favored due to their wide applicability \citep{jacovi-2023-trendsexplainableaixai}. { These methods aim to quantify the importance of each token for a given input and its corresponding model prediction. Such methods can make use of the gradients of the model with respect to its inputs \citep{sundararajan2017axiomatic, simonyan2013deep}, or use surrogate models \citep{ribeiro2016should, lundberg2017unified}.}

    The growing interest in explainable NLP is evidenced by the increasing number of publications surveying explainability in NLP \citep{Liu-xnlp-survey-2019, wallace-xnlp-survey-2020, Zhao-xnlp-survey-2024, madsen-xnlp-survey-2023, zini-xnlp-survey-2020, danilevsky-xnlp-survey-2020}. 
    Additionally, as NLP models are frequently applied in high-stakes domains such as medical \cite{johri-2025-clinical-llm-evaluation} and legal settings \citep{valvoda-towards-xai-legal-prediction-2024} where explainability is essential, a growing number of survey papers now focus on explainability in specific NLP tasks, including fact-checking \citep{kotonya-explainable-fact-checking-survey-2020}, text summarization \citep{dhaini-explainable-summarization-survey-2024}, and for specific explainability methods in NLP \citep{mosca-shap-survey-2022}. Such surveys highlight the wide application of post-hoc methods in NLP. 
    Furthermore, post-hoc methods are used as main explainers in numerous explainability tools and frameworks proposed in the literature \citep{cleverxai-2022, Li-m4-xai-benchmark-2024, agarwal-omni-xai-2022, attanasio-etal-2023-ferret, inseq2023}. These frameworks typically incorporate a range of post-hoc explanation methods while supporting multiple data types and diverse machine learning (ML) model types, including PLMs.
    
    Given the widespread adoption of these methods, evaluating their explanations is increasingly important.  Explanation evaluation has become an active research area in recent years \citep{co12-xai-properties-2023, longo-manifesto-2024}, with numerous metrics and properties proposed \citep{deyoung-etal-2020-eraser, cleverxai-2022, benchmark-BEExAI-2024} to measure the quality of explanations.
    One desirable aspect of an explanation method is subgroup fairness: similar quality of the explanation across subgroups such as the different genders. For example, 
    consider a PLM-based AI system used by clinicians to diagnose patients from textual symptom descriptions and provide post-hoc explanations. The system misdiagnoses both a male and female patient with identical symptoms, where the explanation for the female patient correctly highlights the error, helping the physician identify the mistake. However, the explanation for the male patient falsely emphasizes relevant features in the input text, such as specific symptom-related keywords, misleading the physician into trusting the incorrect diagnosis.
    This discrepancy could undermine trust and harm patient outcomes.

    However, there is a lack of research on evaluating the fairness of explanation methods across demographic groups, particularly 
    in NLP.
   {  
   Most previous works at the intersection of fairness and explainability in NLP explore using explainability as a tool to detect bias in language models \citep{fairnessxai-2024-towards, fairnessxai-2022-challenges, hatexplain-2021, fairness-xai-interplay-2024, inseq2023, vig-gender-bias-plms-2020, gallegos2024bias} or facial recognition models \citep{huber-2023-face-pad-explainability-gender-bias} while some other recent works examine the influence of explanations on human-AI decision-making \citep{explain_HumanAI_2024_1, explain_HumanAI_2024_2}. Despite the rigorous studies on evaluating fairness and bias in language models, less attention has been given to detecting bias in explanations or, in other words, the fairness of explanations themselves.}
    
    In this work,
    we evaluate disparities in the quality of post-hoc explanations across subgroups.  
    We evaluate explanation quality based on a set of key explanation properties. Specifically, we 
    investigate whether explanation methods produce similar faithfulness, robustness, and complexity across demographic groups
    , and focus on gender as a protected attribute. \footnote{Gender, race, age, among others, are referred to as protected attributes under the US anti-discrimination law \citep{xiang-fairness-legal-definitions-2019}}
    We aim to answer the following research question: \textit{Do post-hoc explanation methods perform equivalently across different subgroups, and if not, how can we evaluate gender disparities in explanations? }
    
    \begin{figure*}[t!]
        \centering
        \includegraphics[width=\textwidth]{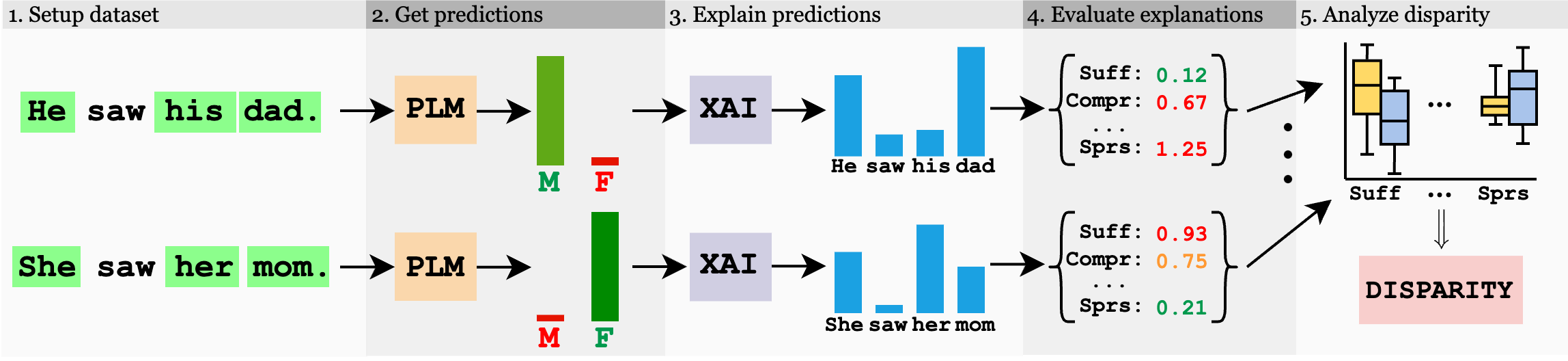}
        \caption{\textbf{Overview of our experimental pipeline}, exemplified with the GECO dataset \citep{wilming2024gecobench}. We begin by obtaining predictions for male/female sentence pairs. We then use feature attribution methods to explain the predictions and evaluate the explanations using various metrics. We finally analyze the distributions of evaluation scores per each metric for male and female sentences and observe if the evaluations differ significantly between the two genders, indicating gender bias and disparity in explanations.}
        \label{fig:pipeline-figure}
    \end{figure*}

    Our findings indicate significant gender disparities in the explanations across different language models, even when the models do not exhibit significant bias.
    Our main contributions are:
    \begin{itemize}
      \item We evaluate the gender disparity in six post-hoc explanation methods on four BERT-based models and GPT-2 using seven evaluation metrics to measure the quality of explanations with respect to their faithfulness, robustness, and complexity. 
      \item We show that all methods can exhibit significant gender disparities regarding all the evaluation metrics used in the experiments.
      \item We further demonstrate that gender disparity in explanations persists even when the models are trained solely on an unbiased dataset, leading to the conclusion that the bias we observe is mainly influenced by the explanation methods.
      \item We finally outline and discuss the implications and considerations for practitioners based on our results.
  \end{itemize}

  We present this work as a step toward raising awareness of gender disparities in explanations and their implications, particularly when interpreting language model outcomes in real-world applications. We hope it contributes to ongoing research efforts aimed at improving the fairness and reliability of post-hoc explainability methods.\footnote{We release our code and datasets on \href{https://github.com/dmah10/fairness-explainable-nlp}{\textbf{GitHub}}} 
  

\section{Related Work}
  
    A number of studies have highlighted limitations of post-hoc explainability methods \citep{xai-new-paradigm-2024, jain-wallace-2019-attention, krishna-disagreement-problem-2024}; however, they fail to consider how these methods perform across different subgroups, thus overlooking issues of fairness of explanations applied to textual datasets. 

    \citet{wilming2024gecobench} study how bias in BERT can influence explanation correctness. They show how re-training and fine-tuning various components of the BERT architecture can improve explanation accuracy in identifying ground-truth tokens. However, this requires a dataset with ground-truth explanations, which is often not the case in more practically relevant datasets. Our study instead evaluates disparities in explanations using multiple metrics that capture three main properties of explanations, none of which requires a dataset with ground-truth explanations. We also introduce a setup designed to minimize any model-induced bias in the explanations, allowing us to investigate gender disparities independent of potential bias in the language models.

    The topic of \textbf{disparities in post-hoc explanations} has been addressed in the literature by three studies, all focusing on tabular datasets \citep{disparity_Dai_2022, fidelity-disparity-2022, understading-disparity-2024}.
    \citet{disparity_Dai_2022} evaluate disparity in the explanation performance with respect to faithfulness (also referred to as fidelity), stability, consistency, and complexity while \citet{fidelity-disparity-2022} focus their evaluation mainly on faithfulness. However, the two works solely experiment on tabular datasets and employ two model classes: linear regression and small neural networks. It remains unclear whether and to what extent explanation methods perform similarly across different subgroups when applied to textual datasets using various PLMs.  Other works explore additional factors that may contribute to the level of disparities exhibited by certain post-hoc explainability methods. \citet{fidelity-disparity-2022} and \citet{understading-disparity-2024} investigate how specific data properties influence disparities in local explanations, with a particular emphasis on faithfulness. In particular, they examine whether the data representation encodes information about the sensitive attribute, and \citet{understading-disparity-2024} further investigates other properties such as limited sample size and covariate shift and evaluates how model characteristics, like model complexity, can result in greater or lesser disparities in the fidelity of LIME \citep{ribeiro2016should} explanations.
     
    \textbf{Textual datasets present unique challenges} compared to tabular datasets. Among the general text-specific challenges in applying explainability methods \citep{zini-xnlp-survey-2020} compared to tabular datasets, isolating sensitive or protected attributes, such as gender,  is particularly more complex in text. This complexity arises from the unstructured nature of text, the implicit representation of gender-related information (as opposed to being explicitly encoded in a single column in tabular datasets), and the context dependency, where gender-related information often depends on the surrounding text. In earlier studies, model selection was limited to either linear regression or a 3- to 4-layer neural network. This work, as detailed later in this paper, explores a diverse set of five transformer-based language models of varying sizes and complexity and two distinct architectures: encoder-only and decoder-only models. Such language models demonstrate high performance in text classification tasks, making them an ideal choice for real-world applications. This underscores the importance of investigating disparities in the post-hoc explanations across subgroups when explanation methods are applied to explain the outcomes of these language models.

    To the best of our knowledge, this paper presents the first study evaluating
    gender disparities in post-hoc explanations with respect to multiple quality metrics on various language models on textual datasets.


\section{Disparity in Post-hoc Explanations}

\subsection{Local Post-hoc Explanation Methods}
\label{sec:post-hoc-methods}


In our evaluation, we focus on six local feature attribution methods: Gradient (Saliency) \cite{simonyan2013deep}, Integrated Gradients (IG) \citep{sundararajan2017axiomatic}, SHAP \citep{lundberg2017unified}, LIME \citep{ribeiro2016should}, and extensions of Gradient and IG in which the input features are multiplied by the importance scores, named Gradient $\times$ Input (GxI) and IG $\times$ Input (IGxI). The applicability of these post-hoc explanation methods has also made them useful for various downstream tasks \citep{zini-xnlp-survey-2020, danilevsky-xnlp-survey-2020}, and also recently for obtaining rationales from smaller models to be used in prompting large language models (LLMs) to improve their performance\citep{amplify-2024, bhan2024self}.


\subsection{Evaluating Explanations} \label{sec:evaluating_explanations}

Prior research on evaluating explanations has introduced various properties and desiderata that can be used to assess the quality of explanation methods and the explanations themselves \citep{explanation-properties-2018, co12-xai-properties-2023}. 
Several studies have built upon these foundational properties to develop metrics for evaluating different aspects of explanation quality. These metrics include both quantitative measures, such as fidelity, stability, consistency, and plausibility \citep{metrics-2021,sensitivity-infidelity-2019,deyoung-etal-2020-eraser,Li-m4-xai-benchmark-2024}, as well as qualitative approaches, which involve human-based evaluations of the generated explanations \citep{qualitative-2020, qualitative-2021} to assess how humans perceive these explanations.

\begin{table*}[h!]
\centering
\small
\caption{\label{tab:metrics}Overview of the considered explanation properties and metrics used to evaluate explanation quality.}
\begin{tabular}{p{2cm}p{3cm}p{9cm}}
\toprule
\textbf{Property} & \textbf{Metric} & \textbf{Definition}   \\ \midrule
\textbf{Faithfulness}  & Comprehensiveness \citep{deyoung-etal-2020-eraser} & Measures whether the explanation captures all the evidence (i.e., tokens) used by the model to make a prediction by assessing the drop in model probability when relevant tokens are removed \\   
 \arrayrulecolor{gray}\cmidrule{2-3}
 & Sufficiency \citep{deyoung-etal-2020-eraser} &  Measures whether the tokens identified by the explanation are sufficient for the model to make a prediction .  \\
 \arrayrulecolor{gray}\cmidrule{2-3}
 & Soft Comprehensiveness \& Soft Sufficiency \citep{aopc-limitations-2023} &  To prevent evaluating explanations on out-of-distribution inputs as a result of removing tokens entirely as in comprehensiveness and sufficiency, for the \textit{soft} versions each token's embedding is masked proportionally to its importance score.  \\
 \arrayrulecolor{black} \midrule
 \textbf{Complexity} & Sparsity \citep{disparity_Dai_2022} &  Counts the number of features with an attributed importance greater than a given threshold.\\
 \arrayrulecolor{gray}\cmidrule{2-3}
 & Gini-index\citep{benchmark-BEExAI-2024} &  Measures the concentration of explanations on specific features by computing the Gini index of attribution vector. A high value, close to 1, indicates a greater concentration of attribution on fewer tokens, which is more desirable compared to a low value, close to 0, where attribution is more evenly distributed across multiple tokens. .    \\
 \arrayrulecolor{black} \midrule
\textbf{Robustness} & Sensitivity \citep{sensitivity-infidelity-2019} & Measures the extent of change in the explanation when there is a slight alteration in the input. High sensitivity in explanations can be problematic, as it may render the explanation method more susceptible to adversarial attacks \cite{Ghorbani_Abid_Zou_2019}. \\
 \arrayrulecolor{black} 
\bottomrule
\end{tabular}
\end{table*} 

In this paper, we quantitatively evaluate explanation quality based on three main desired properties: faithfulness, robustness, and complexity. Table \ref{tab:metrics} presents the metrics we consider to measure the aforementioned properties to evaluate the quality of explanations. 
\subsubsection{Faithfulness} refers to the degree to which an explanation accurately reflects and aligns with the internal workings and decision-making process of a model \citep{jacovi-goldberg-2020-faithfulness}. High faithfulness in explanations is desirable because it ensures that the explanation truly represents the model's functioning in making a prediction. We evaluate faithfulness using four metrics: comprehensiveness, sufficiency,  soft comprehensiveness, and soft sufficiency.
While sufficiency and comprehensiveness are commonly used, recent studies suggest they can lead to inaccurate faithfulness measurements due to the complete token removal operation they use \citep{aopc_unfaithfulness_2022, aopc_unfaithfulness_2022_a}; therefore, we also use soft comprehensiveness and soft sufficiency, which have proven more accurate in measuring faithfulness by masking parts of the tokens' embeddings proportional to their importance scores rather than completely removing a fixed number of tokens \citep{aopc-limitations-2023}. Considering prior literature highlighting disagreement among metrics used to measure faithfulness \citep{jain-wallace-2019-attention, krishna-disagreement-problem-2024}, we employ multiple metrics that differ in evaluating faithfulness. The focus of our study is not to compare these metrics but rather to investigate disparities in explanations with respect to these metrics.

\subsubsection{Robustness}
refers to the degree to which an explainability method responds to small perturbations and changes of the inputs, consistently producing reliable and stable explanations \cite{benchmark-BEExAI-2024, robustness-xai-2018}. In particular, we try to compute the worst-case perturbation that results in the most significant change in the explanations within a region around the original input.
\subsubsection{Complexity}
refers to the degree to which users can easily understand and interpret an explanation. Sparse explanations, compared to dense ones, are generally more favorable as they are less complex and easier to comprehend \citep{benchmark-BEExAI-2024}. We evaluate explanation complexity using two measures: sparsity and Gini index.

We define and provide the \textbf{formulation and the implementation details of these metrics} in Appendix \ref{app:metric-defs}.

\subsection{Implications of Significant Disparity} \label{sec:disparity-implications}

Based on the identified properties, we discuss the implications of the disparity in these properties. Although we focus this paper on gender, the same considerations can apply to other protected attributes.
\textit{Disparities in explanation faithfulness }can result in explanations that do not accurately reflect the model's decision-making process across all groups, potentially leading to less accurate explanations for one group (e.g., female inputs) compared to another (e.g., male inputs). As shown in the example in the introduction, this could undermine stakeholders' trust, leading them to rely on incorrect model outputs.
Significant \textit{disparity in complexity} implies that the explanations for the model's decisions are more complex and, therefore, more challenging to understand for one group compared to another.
\textit{Robustness disparity} implies that explanations for one group exhibit higher sensitivity to slight perturbations, making them more vulnerable to noisy, erroneous data or adversarial attacks.

\section{Experimental Setup}


\subsection{Datasets} \label{sec:dataset}

An ideal dataset to test our hypothesis would contain male and female inputs where the only difference between a pair of male/female inputs is the gender in those inputs, and the difference in gender should have an influence on the task, i.e., the model should not be able to learn to ignore the genders. While ensuring that two inputs differ only in gender is easier to do with tabular data where gender is a categorical attribute, it is harder in textual data in which gender can be apparent in a number of ways: as the subject, or as an object, either explicitly through pronouns or more implicitly through nouns such as \textit{sister/brother} or \textit{actor/actress}. Ensuring that the models cannot ignore the genders is also non-trivial since it is hard to measure what features of a sentence actually reliably influence the inputs. In fact, knowing that would in a way be equivalent to having ground-truth feature importance explanations, as we would know that the words signaling gender in a sentence strongly influence the prediction. {  We experiment with three datasets, taking different approaches with respect to these two considerations. Table \ref{tab:dataset_examples} displays example inputs from each of our datasets.

\begin{table*}[t!]
    \centering
    \small
    \caption{\label{tab:dataset_examples}Example inputs from the datasets used in our experiments. Bold words indicate those that are changed between male and female sentences of each pair.}
    \begin{tabular}{lllll}
        \toprule
        \textbf{Dataset} & \textbf{Examples} & \textbf{Labels} & \textbf{Task} & \textbf{Size}\\
        \midrule
         GECO ALL \citep{wilming2024gecobench}   & 
         \makecell[l]{\textbf{She} is cynically false about \textbf{her} childhood. \\ 
                \textbf{He} is cynically false about \textbf{his} childhood.} & 
         \makecell[l]{Female \\ Male} & 
         Classify gender &
         \makecell[l]{Female: 1,610 \\ Male: 1,610}
         \\
         \midrule
         
         GECO SUBJ \citep{wilming2024gecobench}   & 
         \makecell[l]{ \textbf{She} takes her to a hospital. \\
                \textbf{He} takes her to a hospital.} & 
         \makecell[l]{Female \\ Male} & 
         Classify gender &
         \makecell[l]{Female: 1,610 \\ Male: 1,610}
         \\
         \midrule
         
         Stereotypes   & 
         \makecell[l]{As a \textbf{woman} CFO, \textbf{she} cut budgets ruthlessly. \\ 
                As a \textbf{man} CFO, \textbf{he} cut budgets ruthlessly.} & 
         \makecell[l]{Yes \\ No} & 
         Detect stereotype &
         \makecell[l]{Female: 1,675 \\ Male: 1,675}
         \\
         \midrule
         
         COMPAS \citep{angwin2016compas}   & 
         \makecell[l]{1 priors, score factor 0, under 45, under 25, Hispanic, male \\ 
                0 priors, score factor 0, under 45, under 25, other race, female} & 
         \makecell[l]{Yes \\ No} & 
         Predict recidivism &
         \makecell[l]{Female: 1,175 \\ Male: 4,997}
         \\
        \bottomrule
    \end{tabular}
\end{table*}
    
The first dataset is \textbf{GECO} \citep{wilming2024gecobench}, consisting of pairs of sentences that only differ in their words signaling gender; e.g., replacing \textit{him} with \textit{her} and \textit{sister} with \textit{brother}. The task is to classify the gender in a sentence, either of the entire sentence or only the subject of a sentence. Thus GECO strictly enforces that pairs of sentences are identical except gender, and that those genders strongly influence the predictions, as they \textit{are} the predictions themselves.

Next, inspired by the CrowS-Pairs dataset \citep{nangia-etal-2020-crows}, we construct the synthetic \textbf{Stereotypes} dataset by prompting Claude 3.5 Sonnet \citep{anthropic2024claude} (see Appendix \ref{app:prompts} for the details). It consists of sentence pairs differing only in their gendered words as in GECO, but the task is to classify if a sentence expresses a valid stereotype or not. A valid stereotype is one that is (even if factually inaccurate) associated with one gender more than the other, and the invalid sentences associate the same stereotype with the other gender. This way, we again have pairs of inputs identical except gender, but now the gender in a sentence is not the label directly although it strongly affects it. We validate our dataset first by manually verifying a subset of the inputs, and then by observing that models fine-tuned on this dataset can achieve high accuracy, indicating that the task is meaningful and can be solved with the information in the sentences. 

Finally, we convert the tabular \textbf{COMPAS} \citep{angwin2016compas} dataset for recidivism prediction to text in order to obtain a dataset without pairs identical up to gender, and one in which the gender attribute has a weaker, although not negligible, influence on the task. Following earlier work \citep{fang2024large}, we convert each row to a comma-separated string such as \texttt{3 priors, score factor 1, under 45, under 25, African American, male, misdemeanor}. We do not process the COMPAS dataset to have input pairs that only differ in their gender. That would require assigning labels to previously unseen data points, which we avoid doing to not modify the original relationships between the existing features. 
}

\subsection{Models}

We use five open-source language models that are accessible through the HuggingFace Hub for our experiments, with more information as well as hyperlinks in Table \ref{tab:models} in the appendix. The first two are a base BERT \citep{devlin-etal-2019-bert} model and a distilled TinyBERT \citep{qian2022perturbation}. The third is the GPT-2 model released by OpenAI \citep{radford2019language}, and the fourth is the RoBERTa-large model \citep{roberta-large} which is the largest model we experiment with, with around 355M parameters. Finally, we experiment with a version of BERT released by Meta and named FairBERTa \citep{jiao-etal-2020-tinybert}. We chose to include FairBERTa as it is fine-tuned on a dataset in which inputs containing gender information are perturbed to non-binary words (e.g., he/she $\rightarrow$ they), that is argued to lead to a model which exhibits less disparity between genders. 

\subsection{Explanation Methods \& Evaluation}

For the implementation of our explanation methods (Section~\ref{sec:post-hoc-methods}),
we use the ferret library \citep{attanasio-etal-2023-ferret} which provides off-the-shelf support for models available through the Hugging Face transformers library. We also use ferret and an extension of it\footnote{\url{https://github.com/MatteMartini/Explainable-and-trustworthy-AI-project}} for its implementation of comprehensiveness, sufficiency, and sensitivity metrics. We provide our own implementations based on earlier work for the sparsity, Gini index, and soft sufficiency/comprehensiveness.
Lower values are preferred for sufficiency (AOPC), soft sufficiency, and sparsity, while higher values are preferred for the other metrics.

\subsection{Quantifying Disparity}

We obtain feature-importance explanations for each input in our test set and evaluate the explanations with our metrics, resulting in a list of evaluation scores for the male and female subsets of the dataset, per explanation method and metric. We can then compare these lists per metric to quantify if there is a statistically significant difference or not, and if so how strong it is (i.e. the effect size). 

To measure if the disparity is statistically significant, we follow the previous work \citep{disparity_Dai_2022} and use the Mann-Whitney U test that is applicable to subgroups with different sizes to test the null hypothesis that for any pair of values chosen from the subgroups, they are equally likely to be greater than each other. This corresponds to the methods performing equivalently between the two subgroups. We conclude there is a statistically significant difference if $p \leq 0.05$ and quantify the effect size with the \textit{Cohen's d} metric we define in Appendix \ref{app:cohen}.

\section{Results and Analysis} 



Our main results are shown as follows:
Tables \ref{tab:new_all_table} and \ref{tab:compas_stereo} display the counts of runs (out of five) resulting in statistically significant ($p \leq .05$) disparity, highlighting the cases with considerable effect size ($\vert d \vert \geq 0.2$, following the literature \citep{sawilowsky2009new}) with bold. Moreover, blue cells indicate that the male sentences have higher scores for that metric, while red cells indicate female sentences' scores are higher, with the strength of the color varying with respect to the count in the cell. To evaluate if the disparity we observe is a consequence of the models being pre-trained on biased data, we also report results after training BERT and GPT-2 from scratch on GECO in Section \ref{sec:train_scratch}. We explain our training setup in more detail in Appendix \ref{sec:experimental-pipeline}, and show the average effect sizes of disparities for each configuration in Tables \ref{tab:full_geco_all}, \ref{tab:full_geco_subj}, \ref{tab:full_compas}, \ref{tab:full_stereo} in Appendix \ref{app:additional-results}, as well as further box-plots displaying the distributions of scores in Figures \ref{fig:soft_boxplots} in Appendix \ref{app:additional-results}. We also present a bias analysis in Appendix \ref{sec:bias-analysis} using GECO and show that the models' predictive performance does not exhibit significant disparity.

\begin{table*}
\centering
\caption{\label{tab:new_all_table}\textbf{Number of runs out of five resulting in statistically significant disparity} on the GECO datasets. Cell colors indicate which gender has better evaluation scores for each metric (blue: males, red: females). Bold font further indicates considerable effect size (Cohen's $d$, with $\vert d \vert \geq 0.2$). Metrics are grouped based on the evaluation property they measure.}
\small
\begin{tabular}{clccccccc|ccccccc}
    \toprule
    & & 
    \multicolumn{7}{c|}{GECO-ALL} & \multicolumn{7}{c}{GECO-SUBJ} \\
    \cmidrule{3-16}
    & &
    \multicolumn{4}{c}{Faithf.} & \multicolumn{2}{c}{Comp.} & \multicolumn{1}{c|}{Rbst.} &
     \multicolumn{4}{c}{Faithf.} & \multicolumn{2}{c}{Comp.} & \multicolumn{1}{c}{Rbst.} \\
    \cmidrule{3-16}
    \textbf{Model} & \textbf{Method} & 
    \rot{90}{Compr.} & \rot{90}{Suff.} & \rot{90}{Soft Compr.} & \rot{90}{Soft Suff.} & \rot{90}{Gini} & \rot{90}{Spars.} & \rot{90}{Sens.} & 
    \rot{90}{Compr.} & \rot{90}{Suff.} & \rot{90}{Soft Compr.} & \rot{90}{Soft Suff.} & \rot{90}{Gini} & \rot{90}{Spars.} & \rot{90}{Sens.} \\
    \midrule
     &  Grad &  \textbf{\cellcolor{red!50.0}5} & {\cellcolor{red!50.0}5} & \textbf{\cellcolor{red!50.0}5} & \textbf{\cellcolor{red!50.0}5} & \textbf{\cellcolor{red!40.0}4} & \textbf{\cellcolor{red!30.0}3} & \textbf{\cellcolor{blue!50.0}5} & \textbf{\cellcolor{red!30.0}3} & {\cellcolor{blue!30.0}3} & \textbf{\cellcolor{red!50.0}5} & \textbf{\cellcolor{red!50.0}5} & {\cellcolor{red!30.0}3} & 0 & \textbf{\cellcolor{blue!40.0}4} \\
    &  GxI &  \textbf{\cellcolor{red!40.0}4} & {\cellcolor{red!50.0}5} & \textbf{\cellcolor{red!40.0}4} & \textbf{\cellcolor{red!50.0}5} & \textbf{\cellcolor{red!50.0}5} & \textbf{\cellcolor{red!50.0}5} & \textbf{\cellcolor{blue!50.0}5} & \textbf{\cellcolor{red!50.0}5} & \textbf{\cellcolor{red!30.0}3} & \textbf{\cellcolor{red!50.0}5} & \textbf{\cellcolor{red!50.0}5} & {\cellcolor{red!30.0}3} & 0 & \textbf{\cellcolor{blue!40.0}4} \\
    &  IG &  \textbf{\cellcolor{red!40.0}4} & \textbf{\cellcolor{red!50.0}5} & \textbf{\cellcolor{red!40.0}4} & \textbf{\cellcolor{red!40.0}4} & \textbf{\cellcolor{red!20.0}2} & \textbf{\cellcolor{red!20.0}2} & \textbf{\cellcolor{blue!50.0}5} & \textbf{\cellcolor{red!50.0}5} & \textbf{\cellcolor{blue!40.0}4} & \textbf{\cellcolor{red!50.0}5} & \textbf{\cellcolor{red!50.0}5} & \textbf{\cellcolor{red!40.0}4} & \textbf{\cellcolor{red!40.0}4} & \textbf{\cellcolor{blue!40.0}4} \\
    \multirow[t]{6}{*}{\rotatebox[origin=c]{45}{\textbf{TinyBERT}}} &  IGxI &  \textbf{\cellcolor{red!50.0}5} & {\cellcolor{blue!40.0}4} & \textbf{\cellcolor{red!50.0}5} & \textbf{\cellcolor{red!50.0}5} & \textbf{\cellcolor{red!50.0}5} & \textbf{\cellcolor{red!50.0}5} & \textbf{\cellcolor{blue!50.0}5} & \textbf{\cellcolor{red!30.0}3} & {\cellcolor{blue!50.0}5} & \textbf{\cellcolor{red!50.0}5} & \textbf{\cellcolor{red!50.0}5} & \textbf{\cellcolor{red!50.0}5} & \textbf{\cellcolor{red!50.0}5} & \textbf{\cellcolor{blue!40.0}4} \\
    &  LIME &  \textbf{\cellcolor{red!50.0}5} & {\cellcolor{red!40.0}4} & \textbf{\cellcolor{red!50.0}5} & \textbf{\cellcolor{red!50.0}5} & \textbf{\cellcolor{red!50.0}5} & \textbf{\cellcolor{red!50.0}5} & \textbf{\cellcolor{blue!50.0}5} & \textbf{\cellcolor{red!50.0}5} & {\cellcolor{red!50.0}5} & \textbf{\cellcolor{red!50.0}5} & \textbf{\cellcolor{red!50.0}5} & \textbf{\cellcolor{red!50.0}5} & \textbf{\cellcolor{red!50.0}5} & \textbf{\cellcolor{blue!40.0}4} \\
    &  SHAP &  \textbf{\cellcolor{red!40.0}4} & {\cellcolor{red!30.0}3} & \textbf{\cellcolor{red!50.0}5} & \textbf{\cellcolor{red!50.0}5} & \textbf{\cellcolor{red!50.0}5} & \textbf{\cellcolor{red!50.0}5} & \textbf{\cellcolor{blue!50.0}5} & \textbf{\cellcolor{red!50.0}5} & {\cellcolor{red!50.0}5} & \textbf{\cellcolor{red!50.0}5} & \textbf{\cellcolor{red!50.0}5} & \textbf{\cellcolor{red!50.0}5} & \textbf{\cellcolor{red!50.0}5} & \textbf{\cellcolor{blue!40.0}4} \\
    \cmidrule{1-16}
     &  Grad &  \textbf{\cellcolor{red!50.0}5} & \textbf{\cellcolor{blue!50.0}5} & {\cellcolor{red!50.0}5} & {\cellcolor{red!50.0}5} & {\cellcolor{red!30.0}3} & \textbf{\cellcolor{red!30.0}3} & \textbf{\cellcolor{red!50.0}5} & \textbf{\cellcolor{red!50.0}5} & \textbf{\cellcolor{blue!40.0}4} & \textbf{\cellcolor{red!50.0}5} & \textbf{\cellcolor{red!50.0}5} & \textbf{\cellcolor{blue!20.0}2} & \textbf{\cellcolor{blue!30.0}3} & \textbf{\cellcolor{blue!50.0}5} \\
    &  GxI &  {\cellcolor{red!50.0}5} & \textbf{\cellcolor{blue!50.0}5} & \textbf{\cellcolor{red!40.0}4} & {\cellcolor{red!40.0}4} & \textbf{\cellcolor{blue!50.0}5} & \textbf{\cellcolor{blue!10.0}1} & \textbf{\cellcolor{red!50.0}5} & \textbf{\cellcolor{red!40.0}4} & \textbf{\cellcolor{blue!50.0}5} & \textbf{\cellcolor{red!50.0}5} & \textbf{\cellcolor{red!50.0}5} & \textbf{\cellcolor{blue!20.0}2} & \textbf{\cellcolor{blue!30.0}3} & {\cellcolor{blue!50.0}5} \\
    &  IG &  \textbf{\cellcolor{blue!30.0}3} & \textbf{\cellcolor{blue!50.0}5} & \textbf{\cellcolor{red!40.0}4} & \textbf{\cellcolor{red!50.0}5} & {\cellcolor{blue!10.0}1} & 0 & \textbf{\cellcolor{red!50.0}5} & \textbf{\cellcolor{red!40.0}4} & \textbf{\cellcolor{blue!50.0}5} & \textbf{\cellcolor{red!50.0}5} & \textbf{\cellcolor{red!50.0}5} & {\cellcolor{blue!10.0}1} & 0 & {\cellcolor{red!50.0}5} \\
    \multirow[t]{6}{*}{\rotatebox[origin=c]{45}{\textbf{GPT2}}} &  IGxI &  \textbf{\cellcolor{red!50.0}5} & \textbf{\cellcolor{red!50.0}5} & \textbf{\cellcolor{red!50.0}5} & \textbf{\cellcolor{red!50.0}5} & {\cellcolor{blue!50.0}5} & {\cellcolor{red!40.0}4} & \textbf{\cellcolor{red!50.0}5} & \textbf{\cellcolor{red!50.0}5} & \textbf{\cellcolor{blue!40.0}4} & \textbf{\cellcolor{red!50.0}5} & \textbf{\cellcolor{red!50.0}5} & \textbf{\cellcolor{red!50.0}5} & \textbf{\cellcolor{red!40.0}4} & {\cellcolor{red!50.0}5} \\
    &  LIME &  \textbf{\cellcolor{red!50.0}5} & {\cellcolor{blue!50.0}5} & \textbf{\cellcolor{red!50.0}5} & {\cellcolor{red!50.0}5} & \textbf{\cellcolor{red!50.0}5} & \textbf{\cellcolor{red!20.0}2} & \textbf{\cellcolor{red!50.0}5} & \textbf{\cellcolor{red!40.0}4} & {\cellcolor{red!50.0}5} & \textbf{\cellcolor{red!50.0}5} & \textbf{\cellcolor{red!50.0}5} & {\cellcolor{blue!40.0}4} & {\cellcolor{red!30.0}3} & \textbf{\cellcolor{blue!50.0}5} \\
    &  SHAP &  \textbf{\cellcolor{red!50.0}5} & \textbf{\cellcolor{blue!50.0}5} & {\cellcolor{red!50.0}5} & \textbf{\cellcolor{red!50.0}5} & {\cellcolor{red!50.0}5} & {\cellcolor{red!50.0}5} & \textbf{\cellcolor{red!50.0}5} & \textbf{\cellcolor{red!50.0}5} & {\cellcolor{blue!40.0}4} & \textbf{\cellcolor{red!50.0}5} & \textbf{\cellcolor{red!50.0}5} & \textbf{\cellcolor{red!40.0}4} & \textbf{\cellcolor{red!40.0}4} & \textbf{\cellcolor{blue!50.0}5} \\
    \cmidrule{1-16}
     &  Grad &  \textbf{\cellcolor{red!50.0}5} & {\cellcolor{blue!50.0}5} & \textbf{\cellcolor{red!50.0}5} & \textbf{\cellcolor{red!50.0}5} & {\cellcolor{blue!40.0}4} & 0 & \textbf{\cellcolor{red!50.0}5} & {\cellcolor{red!50.0}5} & {\cellcolor{blue!50.0}5} & \textbf{\cellcolor{red!50.0}5} & \textbf{\cellcolor{red!50.0}5} & \textbf{\cellcolor{blue!50.0}5} & 0 & \textbf{\cellcolor{red!40.0}4} \\
    &  GxI &  \textbf{\cellcolor{red!40.0}4} & \textbf{\cellcolor{blue!40.0}4} & \textbf{\cellcolor{red!50.0}5} & \textbf{\cellcolor{red!50.0}5} & \textbf{\cellcolor{blue!40.0}4} & 0 & \textbf{\cellcolor{red!50.0}5} & \textbf{\cellcolor{red!40.0}4} & \textbf{\cellcolor{blue!20.0}2} & \textbf{\cellcolor{red!50.0}5} & \textbf{\cellcolor{red!50.0}5} & \textbf{\cellcolor{blue!40.0}4} & 0 & \textbf{\cellcolor{red!30.0}3} \\
    &  IG &  \textbf{\cellcolor{red!40.0}4} & \textbf{\cellcolor{blue!40.0}4} & \textbf{\cellcolor{red!40.0}4} & \textbf{\cellcolor{red!40.0}4} & {\cellcolor{red!50.0}5} & \textbf{\cellcolor{red!10.0}1} & \textbf{\cellcolor{red!50.0}5} & {\cellcolor{red!50.0}5} & \textbf{\cellcolor{blue!10.0}1} & \textbf{\cellcolor{red!50.0}5} & \textbf{\cellcolor{red!50.0}5} & {\cellcolor{blue!30.0}3} & 0 & \textbf{\cellcolor{red!40.0}4} \\
    \multirow[t]{6}{*}{\rotatebox[origin=c]{45}{\textbf{BERT}}} &  IGxI &  \textbf{\cellcolor{red!50.0}5} & {\cellcolor{blue!50.0}5} & \textbf{\cellcolor{red!40.0}4} & \textbf{\cellcolor{red!40.0}4} & \textbf{\cellcolor{red!50.0}5} & \textbf{\cellcolor{red!50.0}5} & \textbf{\cellcolor{red!50.0}5} & \textbf{\cellcolor{red!50.0}5} & {\cellcolor{red!40.0}4} & \textbf{\cellcolor{red!50.0}5} & \textbf{\cellcolor{red!50.0}5} & \textbf{\cellcolor{red!50.0}5} & 0 & \textbf{\cellcolor{red!30.0}3} \\
    &  LIME &  \textbf{\cellcolor{red!50.0}5} & {\cellcolor{red!50.0}5} & \textbf{\cellcolor{red!50.0}5} & \textbf{\cellcolor{red!50.0}5} & \textbf{\cellcolor{red!50.0}5} & \textbf{\cellcolor{red!40.0}4} & \textbf{\cellcolor{red!50.0}5} & \textbf{\cellcolor{red!50.0}5} & {\cellcolor{red!30.0}3} & \textbf{\cellcolor{red!50.0}5} & \textbf{\cellcolor{red!50.0}5} & \textbf{\cellcolor{red!30.0}3} & \textbf{\cellcolor{red!30.0}3} & \textbf{\cellcolor{red!40.0}4} \\
    &  SHAP &  \textbf{\cellcolor{red!50.0}5} & {\cellcolor{blue!50.0}5} & \textbf{\cellcolor{red!40.0}4} & \textbf{\cellcolor{red!50.0}5} & \textbf{\cellcolor{red!50.0}5} & \textbf{\cellcolor{red!50.0}5} & \textbf{\cellcolor{red!50.0}5} & \textbf{\cellcolor{red!50.0}5} & {\cellcolor{red!20.0}2} & \textbf{\cellcolor{red!50.0}5} & \textbf{\cellcolor{red!50.0}5} & {\cellcolor{red!40.0}4} & {\cellcolor{red!40.0}4} & \textbf{\cellcolor{red!40.0}4} \\
    \cmidrule{1-16}
     &  Grad &  \textbf{\cellcolor{red!50.0}5} & \textbf{\cellcolor{blue!40.0}4} & \textbf{\cellcolor{red!50.0}5} & \textbf{\cellcolor{red!50.0}5} & \textbf{\cellcolor{blue!40.0}4} & 0 & \textbf{\cellcolor{red!40.0}4} & \textbf{\cellcolor{red!30.0}3} & {\cellcolor{red!50.0}5} & \textbf{\cellcolor{blue!50.0}5} & \textbf{\cellcolor{blue!50.0}5} & {\cellcolor{blue!40.0}4} & 0 & \textbf{\cellcolor{red!40.0}4} \\
    &  GxI &  \textbf{\cellcolor{red!30.0}3} & \textbf{\cellcolor{blue!40.0}4} & \textbf{\cellcolor{red!50.0}5} & \textbf{\cellcolor{red!50.0}5} & \textbf{\cellcolor{blue!20.0}2} & 0 & \textbf{\cellcolor{red!40.0}4} & \textbf{\cellcolor{red!30.0}3} & {\cellcolor{red!30.0}3} & {\cellcolor{blue!50.0}5} & {\cellcolor{blue!50.0}5} & {\cellcolor{red!30.0}3} & {\cellcolor{red!10.0}1} & \textbf{\cellcolor{red!40.0}4} \\
    &  IG &  \textbf{\cellcolor{red!30.0}3} & \textbf{\cellcolor{blue!40.0}4} & \textbf{\cellcolor{red!50.0}5} & \textbf{\cellcolor{red!50.0}5} & \textbf{\cellcolor{red!40.0}4} & \textbf{\cellcolor{red!30.0}3} & \textbf{\cellcolor{red!40.0}4} & {\cellcolor{red!40.0}4} & \textbf{\cellcolor{blue!30.0}3} & {\cellcolor{blue!50.0}5} & {\cellcolor{blue!50.0}5} & \textbf{\cellcolor{red!20.0}2} & {\cellcolor{red!20.0}2} & \textbf{\cellcolor{red!40.0}4} \\
    \multirow[t]{6}{*}{\rotatebox[origin=c]{45}{\textbf{FairBERTa}}} &  IGxI &  \textbf{\cellcolor{red!50.0}5} & \textbf{\cellcolor{blue!40.0}4} & \textbf{\cellcolor{red!50.0}5} & \textbf{\cellcolor{red!50.0}5} & \textbf{\cellcolor{red!40.0}4} & \textbf{\cellcolor{red!40.0}4} & \textbf{\cellcolor{red!40.0}4} & \textbf{\cellcolor{red!40.0}4} & {\cellcolor{red!30.0}3} & {\cellcolor{blue!50.0}5} & {\cellcolor{blue!50.0}5} & \textbf{\cellcolor{red!50.0}5} & \textbf{\cellcolor{red!50.0}5} & \textbf{\cellcolor{red!40.0}4} \\
    &  LIME &  \textbf{\cellcolor{red!50.0}5} & \textbf{\cellcolor{blue!40.0}4} & \textbf{\cellcolor{red!50.0}5} & \textbf{\cellcolor{red!50.0}5} & \textbf{\cellcolor{red!30.0}3} & \textbf{\cellcolor{red!30.0}3} & \textbf{\cellcolor{red!40.0}4} & \textbf{\cellcolor{red!50.0}5} & {\cellcolor{red!30.0}3} & {\cellcolor{blue!50.0}5} & {\cellcolor{blue!50.0}5} & \textbf{\cellcolor{red!40.0}4} & \textbf{\cellcolor{red!40.0}4} & \textbf{\cellcolor{red!40.0}4} \\
    &  SHAP &  \textbf{\cellcolor{red!50.0}5} & \textbf{\cellcolor{blue!40.0}4} & \textbf{\cellcolor{red!50.0}5} & \textbf{\cellcolor{red!50.0}5} & \textbf{\cellcolor{red!30.0}3} & \textbf{\cellcolor{red!30.0}3} & \textbf{\cellcolor{red!40.0}4} & \textbf{\cellcolor{red!50.0}5} & {\cellcolor{red!40.0}4} & {\cellcolor{blue!50.0}5} & {\cellcolor{blue!50.0}5} & \textbf{\cellcolor{red!30.0}3} & \textbf{\cellcolor{red!30.0}3} & \textbf{\cellcolor{red!40.0}4} \\
    \cmidrule{1-16}
     &  Grad &  \textbf{\cellcolor{red!50.0}5} & \textbf{\cellcolor{blue!50.0}5} & \textbf{\cellcolor{red!50.0}5} & \textbf{\cellcolor{red!50.0}5} & 0 & 0 & \textbf{\cellcolor{red!50.0}5} & \textbf{\cellcolor{red!50.0}5} & \textbf{\cellcolor{blue!50.0}5} & \textbf{\cellcolor{red!50.0}5} & \textbf{\cellcolor{red!50.0}5} & 0 & 0 & \textbf{\cellcolor{red!50.0}5} \\
    &  GxI &  \textbf{\cellcolor{red!20.0}2} & \textbf{\cellcolor{blue!50.0}5} & \textbf{\cellcolor{red!50.0}5} & \textbf{\cellcolor{red!50.0}5} & {\cellcolor{red!20.0}2} & {\cellcolor{red!10.0}1} & \textbf{\cellcolor{red!50.0}5} & \textbf{\cellcolor{red!30.0}3} & \textbf{\cellcolor{blue!50.0}5} & \textbf{\cellcolor{red!50.0}5} & \textbf{\cellcolor{red!50.0}5} & 0 & 0 & \textbf{\cellcolor{red!50.0}5} \\
    &  IG &  {\cellcolor{red!30.0}3} & \textbf{\cellcolor{blue!50.0}5} & \textbf{\cellcolor{red!50.0}5} & \textbf{\cellcolor{red!50.0}5} & 0 & 0 & {\cellcolor{blue!50.0}5} & \textbf{\cellcolor{red!10.0}1} & \textbf{\cellcolor{blue!50.0}5} & \textbf{\cellcolor{red!50.0}5} & \textbf{\cellcolor{red!50.0}5} & 0 & 0 & \textbf{\cellcolor{red!50.0}5} \\
    \multirow[t]{6}{*}{\rotatebox[origin=c]{45}{\textbf{RoBERTa}}} &  IGxI &  \textbf{\cellcolor{red!40.0}4} & \textbf{\cellcolor{blue!50.0}5} & \textbf{\cellcolor{red!50.0}5} & \textbf{\cellcolor{red!50.0}5} & \textbf{\cellcolor{red!10.0}1} & 0 & \textbf{\cellcolor{red!50.0}5} & \textbf{\cellcolor{red!40.0}4} & \textbf{\cellcolor{blue!50.0}5} & \textbf{\cellcolor{red!50.0}5} & \textbf{\cellcolor{red!50.0}5} & {\cellcolor{red!20.0}2} & {\cellcolor{red!20.0}2} & \textbf{\cellcolor{red!50.0}5} \\
    &  LIME &  \textbf{\cellcolor{red!50.0}5} & \textbf{\cellcolor{blue!50.0}5} & \textbf{\cellcolor{red!50.0}5} & \textbf{\cellcolor{red!50.0}5} & \textbf{\cellcolor{red!50.0}5} & \textbf{\cellcolor{red!50.0}5} & \textbf{\cellcolor{red!50.0}5} & \textbf{\cellcolor{red!50.0}5} & \textbf{\cellcolor{blue!40.0}4} & \textbf{\cellcolor{red!50.0}5} & \textbf{\cellcolor{red!50.0}5} & \textbf{\cellcolor{red!50.0}5} & \textbf{\cellcolor{red!50.0}5} & \textbf{\cellcolor{red!50.0}5} \\
    &  SHAP &  \textbf{\cellcolor{red!50.0}5} & \textbf{\cellcolor{blue!50.0}5} & \textbf{\cellcolor{red!40.0}4} & \textbf{\cellcolor{red!40.0}4} & \textbf{\cellcolor{red!50.0}5} & \textbf{\cellcolor{red!50.0}5} & \textbf{\cellcolor{red!50.0}5} & \textbf{\cellcolor{red!50.0}5} & \textbf{\cellcolor{blue!50.0}5} & \textbf{\cellcolor{red!50.0}5} & \textbf{\cellcolor{red!50.0}5} & \textbf{\cellcolor{red!50.0}5} & \textbf{\cellcolor{red!40.0}4} & \textbf{\cellcolor{red!50.0}5} \\
    \bottomrule
\end{tabular}
\end{table*}

\begin{table*}
\centering
\caption{\label{tab:compas_stereo}\textbf{Number of runs out of five resulting in statistically significant disparity} on the COMPAS and Stereotypes datasets. Cell colors indicate which gender has better evaluation scores for each metric (blue: males, red: females). Bold font further indicates considerable effect size (Cohen's $d$, with $\vert d \vert \geq 0.2$). Metrics are grouped based on the evaluation property they measure.}
\small
\begin{tabular}{clrrrrrrr|rrrrrrr}
    \toprule
    & & \multicolumn{7}{c|}{COMPAS} & \multicolumn{7}{c}{Stereotypes} \\
    \cmidrule{3-16}

      & &
    \multicolumn{4}{c}{Faithf.} & \multicolumn{2}{c}{Comp.} & \multicolumn{1}{c|}{Rbst.} &
     \multicolumn{4}{c}{Faithf.} & \multicolumn{2}{c}{Comp.} & \multicolumn{1}{c}{Rbst.} \\
    \cmidrule{3-16}
    
    \textbf{Model} & \textbf{Method} & 
    \rot{90}{Compr.} & \rot{90}{Suff.} & \rot{90}{Soft Compr.} & \rot{90}{Soft Suff.} & \rot{90}{Gini} & \rot{90}{Spars.} & \rot{90}{Sens.} &
    \rot{90}{Compr.} & \rot{90}{Suff.} & \rot{90}{Soft Compr.} & \rot{90}{Soft Suff.} & \rot{90}{Gini} & \rot{90}{Spars.} & \rot{90}{Sens.} \\
    \midrule
    
    & Grad & \textbf{\cellcolor{blue!40.0}4} & {\cellcolor{blue!40.0}4} & \textbf{\cellcolor{red!10.0}1} & \textbf{\cellcolor{red!10.0}1} & \textbf{\cellcolor{blue!20.0}2} & {\cellcolor{red!50.0}5} & \textbf{\cellcolor{blue!40.0}4} &
    \textbf{\cellcolor{blue!50.0}5} & \textbf{\cellcolor{red!50.0}5} & 0 & {\cellcolor{red!50.0}5} & \textbf{\cellcolor{blue!50.0}5} & 0 & \textbf{\cellcolor{blue!50.0}5} \\
    
    & GxI & \textbf{\cellcolor{blue!50.0}5} & \textbf{\cellcolor{blue!20.0}2} & \textbf{\cellcolor{red!10.0}1} & \textbf{\cellcolor{red!10.0}1} & \textbf{\cellcolor{blue!50.0}5} & \textbf{\cellcolor{blue!50.0}5} & \textbf{\cellcolor{blue!40.0}4} &
    0 & \textbf{\cellcolor{red!50.0}5} & 0 & 0 & \textbf{\cellcolor{blue!50.0}5} & 0 & \textbf{\cellcolor{blue!50.0}5} \\
    
    \multirow[t]{4}{*}{\rotatebox[origin=c]{45}{\textbf{TinyBERT}}} & 
    IG & \textbf{\cellcolor{blue!10.0}1} & {\cellcolor{red!50.0}5} & \textbf{\cellcolor{red!10.0}1} & \textbf{\cellcolor{red!10.0}1} & {\cellcolor{blue!20.0}2} & {\cellcolor{red!20.0}2} & \textbf{\cellcolor{blue!50.0}5} &
    0 & \textbf{\cellcolor{red!50.0}5} & 0 & 0 & \textbf{\cellcolor{blue!50.0}5} & \textbf{\cellcolor{blue!50.0}5} & \textbf{\cellcolor{blue!50.0}5} \\
    
    & IGxI & \textbf{\cellcolor{blue!40.0}4} & {\cellcolor{red!40.0}4} & \textbf{\cellcolor{red!10.0}1} & \textbf{\cellcolor{red!10.0}1} & \textbf{\cellcolor{blue!50.0}5} & \textbf{\cellcolor{blue!40.0}4} & \textbf{\cellcolor{blue!50.0}5} &
    \textbf{\cellcolor{red!50.0}5} & \textbf{\cellcolor{red!50.0}5} & 0 & 0 & \textbf{\cellcolor{red!50.0}5} & \textbf{\cellcolor{red!50.0}5} & \textbf{\cellcolor{blue!50.0}5} \\
    
    & LIME & {\cellcolor{blue!50.0}5} & {\cellcolor{red!40.0}4} & \textbf{\cellcolor{red!10.0}1} & \textbf{\cellcolor{red!10.0}1} & \textbf{\cellcolor{blue!40.0}4} & \textbf{\cellcolor{blue!10.0}1} & \textbf{\cellcolor{blue!30.0}3} &
    {\cellcolor{red!50.0}5} & \textbf{\cellcolor{red!50.0}5} & 0 & 0 & \textbf{\cellcolor{red!50.0}5} & \textbf{\cellcolor{red!50.0}5} & \textbf{\cellcolor{blue!50.0}5} \\
    
    & SHAP & \textbf{\cellcolor{blue!30.0}3} & {\cellcolor{red!40.0}4} & \textbf{\cellcolor{red!10.0}1} & \textbf{\cellcolor{red!10.0}1} & \textbf{\cellcolor{blue!40.0}4} & {\cellcolor{blue!50.0}5} & \textbf{\cellcolor{blue!50.0}5} &
    \textbf{\cellcolor{red!50.0}5} & \textbf{\cellcolor{red!50.0}5} & 0 & 0 & {\cellcolor{red!50.0}5} & 0 & \textbf{\cellcolor{blue!50.0}5} \\
    \cmidrule{1-16}
    
    & Grad & \textbf{\cellcolor{red!50.0}5} & \textbf{\cellcolor{red!40.0}4} & \textbf{\cellcolor{red!10.0}1} & {\cellcolor{blue!40.0}4} & {\cellcolor{blue!30.0}3} & \textbf{\cellcolor{red!10.0}1} & \textbf{\cellcolor{blue!40.0}4} &
    \textbf{\cellcolor{blue!30.0}3} & {\cellcolor{red!30.0}3} & {\cellcolor{red!20.0}2} & {\cellcolor{red!20.0}2} & \textbf{\cellcolor{blue!50.0}5} & 0 & \textbf{\cellcolor{blue!30.0}3} \\
    
    & GxI & \textbf{\cellcolor{red!50.0}5} & \textbf{\cellcolor{blue!40.0}4} & {\cellcolor{blue!30.0}3} & \textbf{\cellcolor{blue!20.0}2} & \textbf{\cellcolor{red!50.0}5} & {\cellcolor{blue!40.0}4} & \textbf{\cellcolor{blue!40.0}4} &
    \textbf{\cellcolor{blue!20.0}2} & {\cellcolor{blue!50.0}5} & \textbf{\cellcolor{red!20.0}2} & \textbf{\cellcolor{red!20.0}2} & \textbf{\cellcolor{blue!50.0}5} & 0 & {\cellcolor{blue!40.0}4} \\
    
    \multirow[t]{4}{*}{\rotatebox[origin=c]{45}{\textbf{GPT2}}} &
    IG & {\cellcolor{red!40.0}4} & \textbf{\cellcolor{blue!40.0}4} & {\cellcolor{blue!50.0}5} & {\cellcolor{blue!50.0}5} & \textbf{\cellcolor{blue!10.0}1} & {\cellcolor{blue!40.0}4} & \textbf{\cellcolor{blue!30.0}3} &
    0 & \textbf{\cellcolor{red!30.0}3} & 0 & \textbf{\cellcolor{red!20.0}2} & \textbf{\cellcolor{red!30.0}3} & 0 & {\cellcolor{blue!20.0}2} \\
    
    & IGxI & \textbf{\cellcolor{red!40.0}4} & {\cellcolor{red!30.0}3} & {\cellcolor{blue!50.0}5} & {\cellcolor{blue!50.0}5} & \textbf{\cellcolor{red!50.0}5} & \textbf{\cellcolor{red!40.0}4} & \textbf{\cellcolor{blue!10.0}1} &
    {\cellcolor{blue!30.0}3} & \textbf{\cellcolor{red!50.0}5} & 0 & \textbf{\cellcolor{red!20.0}2} & \textbf{\cellcolor{red!20.0}2} & {\cellcolor{blue!50.0}5} & {\cellcolor{red!10.0}1} \\
    
    & LIME & \textbf{\cellcolor{red!50.0}5} & \textbf{\cellcolor{blue!20.0}2} & \textbf{\cellcolor{blue!10.0}1} & {\cellcolor{blue!40.0}4} & \textbf{\cellcolor{blue!40.0}4} & \textbf{\cellcolor{blue!30.0}3} & \textbf{\cellcolor{blue!40.0}4} &
    \textbf{\cellcolor{red!30.0}3} & \textbf{\cellcolor{red!30.0}3} & \textbf{\cellcolor{red!20.0}2} & {\cellcolor{red!20.0}2} & \textbf{\cellcolor{red!30.0}3} & {\cellcolor{red!30.0}3} & \textbf{\cellcolor{blue!30.0}3} \\
    
    & SHAP & \textbf{\cellcolor{red!40.0}4} & {\cellcolor{blue!30.0}3} & 0 & 0 & \textbf{\cellcolor{blue!50.0}5} & \textbf{\cellcolor{blue!40.0}4} & \textbf{\cellcolor{blue!40.0}4} &
    \textbf{\cellcolor{red!50.0}5} & \textbf{\cellcolor{red!50.0}5} & 0 & 0 & \textbf{\cellcolor{red!50.0}5} & \textbf{\cellcolor{red!30.0}3} & \textbf{\cellcolor{blue!20.0}2} \\
    \cmidrule{1-16}
    
    & Grad & \textbf{\cellcolor{red!50.0}5} & {\cellcolor{blue!20.0}2} & \textbf{\cellcolor{red!30.0}3} & \textbf{\cellcolor{red!40.0}4} & {\cellcolor{blue!30.0}3} & \textbf{\cellcolor{red!50.0}5} & \textbf{\cellcolor{blue!40.0}4} &
    \textbf{\cellcolor{blue!50.0}5} & {\cellcolor{blue!50.0}5} & 0 & 0 & \textbf{\cellcolor{blue!50.0}5} & 0 & \textbf{\cellcolor{red!30.0}3} \\
    
    & GxI & \textbf{\cellcolor{red!50.0}5} & {\cellcolor{red!30.0}3} & \textbf{\cellcolor{red!30.0}3} & \textbf{\cellcolor{red!30.0}3} & \textbf{\cellcolor{red!30.0}3} & \textbf{\cellcolor{red!30.0}3} & \textbf{\cellcolor{blue!20.0}2} &
    \textbf{\cellcolor{blue!20.0}2} & \textbf{\cellcolor{red!20.0}2} & 0 & 0 & \textbf{\cellcolor{blue!30.0}3} & \textbf{\cellcolor{blue!50.0}5} & \textbf{\cellcolor{red!20.0}2} \\
    
    \multirow[t]{4}{*}{\rotatebox[origin=c]{45}{\textbf{BERT}}} &
    IG & \textbf{\cellcolor{red!20.0}2} & \textbf{\cellcolor{blue!10.0}1} & \textbf{\cellcolor{red!30.0}3} & \textbf{\cellcolor{red!30.0}3} & {\cellcolor{red!40.0}4} & \textbf{\cellcolor{red!30.0}3} & \textbf{\cellcolor{blue!30.0}3} &
    \textbf{\cellcolor{blue!50.0}5} & \textbf{\cellcolor{red!50.0}5} & 0 & 0 & 0 & 0 & \textbf{\cellcolor{red!40.0}4} \\
    
    & IGxI & \textbf{\cellcolor{red!50.0}5} & \textbf{\cellcolor{blue!50.0}5} & \textbf{\cellcolor{red!30.0}3} & \textbf{\cellcolor{red!40.0}4} & {\cellcolor{blue!20.0}2} & {\cellcolor{red!40.0}4} & \textbf{\cellcolor{blue!30.0}3} &
    \textbf{\cellcolor{blue!20.0}2} & \textbf{\cellcolor{red!50.0}5} & 0 & 0 & 0 & 0 & \textbf{\cellcolor{red!40.0}4} \\
    
    & LIME & \textbf{\cellcolor{red!50.0}5} & \textbf{\cellcolor{blue!40.0}4} & \textbf{\cellcolor{red!30.0}3} & \textbf{\cellcolor{red!30.0}3} & \textbf{\cellcolor{blue!30.0}3} & {\cellcolor{red!20.0}2} & \textbf{\cellcolor{blue!30.0}3} &
    \textbf{\cellcolor{blue!20.0}2} & \textbf{\cellcolor{red!50.0}5} & 0 & 0 & \textbf{\cellcolor{blue!50.0}5} & \textbf{\cellcolor{blue!20.0}2} & \textbf{\cellcolor{red!20.0}2} \\
    
    & SHAP & \textbf{\cellcolor{red!50.0}5} & \textbf{\cellcolor{blue!40.0}4} & \textbf{\cellcolor{red!20.0}2} & \textbf{\cellcolor{red!30.0}3} & {\cellcolor{red!40.0}4} & \textbf{\cellcolor{red!50.0}5} & \textbf{\cellcolor{blue!30.0}3} &
    \textbf{\cellcolor{blue!50.0}5} & {\cellcolor{blue!50.0}5} & 0 & 0 & \textbf{\cellcolor{blue!20.0}2} & \textbf{\cellcolor{blue!50.0}5} & {\cellcolor{red!50.0}5} \\
    \cmidrule{1-16}
    
    & Grad & \textbf{\cellcolor{red!20.0}2} & \textbf{\cellcolor{red!10.0}1} & {\cellcolor{red!20.0}2} & {\cellcolor{red!20.0}2} & {\cellcolor{blue!40.0}4} & 0 & {\cellcolor{blue!40.0}4} &
    0 & 0 & 0 & 0 & \textbf{\cellcolor{red!50.0}5} & 0 & \textbf{\cellcolor{blue!10.0}1} \\
    
    & GxI & \textbf{\cellcolor{red!20.0}2} & {\cellcolor{red!30.0}3} & {\cellcolor{red!20.0}2} & \textbf{\cellcolor{red!10.0}1} & \textbf{\cellcolor{blue!40.0}4} & {\cellcolor{blue!30.0}3} & {\cellcolor{red!40.0}4} &
    0 & \textbf{\cellcolor{red!50.0}5} & 0 & 0 & 0 & 0 & \textbf{\cellcolor{blue!10.0}1} \\
    
    \multirow[t]{4}{*}{\rotatebox[origin=c]{45}{\textbf{FairBERTa}}} & 
    IG & {\cellcolor{blue!50.0}5} & \textbf{\cellcolor{blue!40.0}4} & {\cellcolor{red!20.0}2} & {\cellcolor{red!20.0}2} & {\cellcolor{blue!40.0}4} & {\cellcolor{red!20.0}2} & {\cellcolor{blue!20.0}2} &
    0 & \textbf{\cellcolor{red!50.0}5} & 0 & 0 & 0 & 0 & \textbf{\cellcolor{blue!30.0}3} \\
    
    & IGxI & {\cellcolor{blue!30.0}3} & \textbf{\cellcolor{blue!20.0}2} & \textbf{\cellcolor{red!10.0}1} & {\cellcolor{red!20.0}2} & {\cellcolor{blue!20.0}2} & {\cellcolor{red!40.0}4} & \textbf{\cellcolor{blue!30.0}3} &
    0 & \textbf{\cellcolor{red!50.0}5} & 0 & 0 & \textbf{\cellcolor{blue!50.0}5} & \textbf{\cellcolor{blue!50.0}5} & {\cellcolor{blue!10.0}1} \\
    
    & LIME & \textbf{\cellcolor{red!10.0}1} & \textbf{\cellcolor{blue!10.0}1} & \textbf{\cellcolor{red!10.0}1} & \textbf{\cellcolor{red!10.0}1} & {\cellcolor{blue!50.0}5} & {\cellcolor{blue!20.0}2} & {\cellcolor{red!30.0}3} &
    \textbf{\cellcolor{blue!50.0}5} & \textbf{\cellcolor{blue!50.0}5} & 0 & 0 & 0 & 0 & \textbf{\cellcolor{blue!10.0}1} \\
    
    & SHAP & \textbf{\cellcolor{red!10.0}1} & \textbf{\cellcolor{blue!10.0}1} & {\cellcolor{red!20.0}2} & {\cellcolor{red!20.0}2} & {\cellcolor{blue!40.0}4} & {\cellcolor{blue!20.0}2} & {\cellcolor{red!30.0}3} &
    \textbf{\cellcolor{red!50.0}5} & {\cellcolor{blue!50.0}5} & 0 & 0 & \textbf{\cellcolor{blue!50.0}5} & \textbf{\cellcolor{blue!50.0}5} & \textbf{\cellcolor{blue!30.0}3} \\
    \cmidrule{1-16}
    
    & Grad & {\cellcolor{blue!30.0}3} & {\cellcolor{blue!20.0}2} & \textbf{\cellcolor{red!10.0}1} & \textbf{\cellcolor{red!20.0}2} & {\cellcolor{red!30.0}3} & 0 & 0 &
    {\cellcolor{blue!50.0}5} & \textbf{\cellcolor{red!20.0}2} & {\cellcolor{red!40.0}4} & {\cellcolor{red!40.0}4} & {\cellcolor{red!40.0}4} & 0 & 0 \\
    
    & GxI & {\cellcolor{blue!10.0}1} & {\cellcolor{blue!20.0}2} & {\cellcolor{red!20.0}2} & \textbf{\cellcolor{red!10.0}1} & \textbf{\cellcolor{blue!10.0}1} & \textbf{\cellcolor{blue!10.0}1} & 0 &
    {\cellcolor{blue!10.0}1} & \textbf{\cellcolor{red!20.0}2} & {\cellcolor{blue!10.0}1} & {\cellcolor{blue!10.0}1} & {\cellcolor{blue!30.0}3} & {\cellcolor{red!10.0}1} & 0 \\
    
    \multirow[t]{4}{*}{\rotatebox[origin=c]{45}{\textbf{RoBERTa}}} & 
    IG & \textbf{\cellcolor{red!10.0}1} & {\cellcolor{red!30.0}3} & \textbf{\cellcolor{red!10.0}1} & \textbf{\cellcolor{red!10.0}1} & \textbf{\cellcolor{red!10.0}1} & \textbf{\cellcolor{blue!10.0}1} & 0 &
    \textbf{\cellcolor{blue!20.0}2} & {\cellcolor{red!40.0}4} & {\cellcolor{red!10.0}1} & {\cellcolor{red!30.0}3} & \textbf{\cellcolor{blue!10.0}1} & {\cellcolor{blue!20.0}2} & 0 \\
    
    & IGxI & {\cellcolor{blue!30.0}3} & \textbf{\cellcolor{red!30.0}3} & \textbf{\cellcolor{red!10.0}1} & {\cellcolor{red!20.0}2} & \textbf{\cellcolor{red!30.0}3} & {\cellcolor{blue!30.0}3} & 0 &
    \textbf{\cellcolor{blue!30.0}3} & {\cellcolor{red!40.0}4} & {\cellcolor{blue!10.0}1} & {\cellcolor{blue!10.0}1} & \textbf{\cellcolor{blue!20.0}2} & \textbf{\cellcolor{blue!10.0}1} & 0 \\
    
    & LIME & \textbf{\cellcolor{red!10.0}1} & {\cellcolor{red!10.0}1} & \textbf{\cellcolor{red!10.0}1} & \textbf{\cellcolor{red!10.0}1} & {\cellcolor{red!10.0}1} & 0 & 0 &
    {\cellcolor{blue!50.0}5} & {\cellcolor{red!40.0}4} & {\cellcolor{red!40.0}4} & {\cellcolor{red!40.0}4} & {\cellcolor{red!40.0}4} & {\cellcolor{red!40.0}4} & {\cellcolor{red!30.0}3} \\
    
    & SHAP & {\cellcolor{blue!20.0}2} & {\cellcolor{red!30.0}3} & \textbf{\cellcolor{red!10.0}1} & \textbf{\cellcolor{red!10.0}1} & \textbf{\cellcolor{blue!20.0}2} & {\cellcolor{blue!40.0}4} & 0 &
    {\cellcolor{blue!40.0}4} & {\cellcolor{red!50.0}5} & {\cellcolor{blue!40.0}4} & {\cellcolor{red!10.0}1} & {\cellcolor{blue!50.0}5} & \textbf{\cellcolor{blue!40.0}4} & \textbf{\cellcolor{red!10.0}1} \\
    \bottomrule
\end{tabular}
\end{table*}

In total, of the 5,040 combinations of dataset, model, explanation, metric, seeds in our experiments, 3,647 (72.4\%) exhibit statistically significant ($p \leq 0.05$) disparity and 2,761 (54.8\%) do so with a considerable effect size ($\vert d \vert \geq 0.2$).

\subsection{Disparity per Explanation Method}

To analyze disparity per method, we aggregate results in Tables \ref{tab:new_all_table},\ref{tab:compas_stereo}, and \ref{tab:random_init} (or Tables \ref{tab:full_geco_all}, \ref{tab:full_geco_subj},\ref{tab:full_compas} and \ref{tab:full_stereo} in the Appendix) row-wise considering all experimental combinations with metrics, models and datasets.
Results show that  IGxI (60\%), SHAP (59\%) and LIME (57\%) exhibit the highest values
for significant disparity with considerable effect size 
followed by Grad (51.3\%), GxI (51.1\%), and IG (49\%) where these values are notably high and reflect significant disparity and bias in the performance of these explanations.
Excluding IGxI, gradient-based methods (Grad, GxI, IG) show relatively less significant explanation disparity than perturbation-based methods (SHAP and LIME). 
Overall, results demonstrate that all six methods we considered exhibit significant disparity with considerable effect size on more than 49\% of the combinations. These results are also reflected in the differences and gaps between the evaluation scores distributions in Figure \ref{fig:boxplots} (and Figure \ref{fig:soft_boxplots} in Appendix \ref{app:additional-results}).



\subsection{Disparity Across Metrics}



\subsubsection{Faithfulness Disparity}

On both GECO datasets, more than 95\% of the runs with \textit{soft sufficiency} and \textit{soft comprehensiveness} results in significant disparity, with 85\% also exhibiting considerable effect size, while the \textit{comprehensiveness} and \textit{sufficiency} less frequently result in significant disparity. Most noticeably on GECO-SUBJ, all runs with the soft metrics result in significant disparity. Furthermore, for all faithfulness metrics, the direction of their disparities for each model is often consistent between the explanation methods, indicating that the model plays a larger role in determining this direction than the explanation method.

In particular, the soft metrics show a noticeable decrease in the number of runs with significant disparity on the COMPAS and Stereotypes datasets, to less than 40\% on COMPAS and 20\% on Stereotypes. Nevertheless, the regular comprehensiveness and sufficiency metrics show a smaller decrease with 63\% of runs on COMPAS and 71.5\% on Stereotypes showing significant disparity. These results indicate that the soft removal operations can help reduce disparities in the faithfulness of explanations as long as the sensitive attribute is not the label directly.

We also observe in particular on the more practically relevant COMPAS dataset that unbiased pre-training as in FairBERTa and using larger models such as RoBERTa might help reduce the occurrence of disparities. Nevertheless, the high amount of faithfulness disparities visible across models and explanation methods for the comprehensiveness and sufficiency metrics highlights that feature attribution methods can lead to unfair performance between sensitive attributes such as gender.
\subsubsection{Complexity Disparity}

On the GECO datasets, the complexity metrics Gini index and \textit{sparsity}, exhibit disparity less frequently than the faithfulness metrics, at 70\% and 49.5\% respectively, with 53\% and 42\% of the total runs also resulting in disparity with considerable effect size. Results for the COMPAS and Stereotypes datasets also follow similar percentages, with 65\% (38\% with considerable effect size) and 68\% (54\% with considerable effect size) for the Gini Index on COMPAS and Stereotypes, and likewise 57\% (27\%) and 40\% (30\%) for sparsity. 

Similar to the results with faithfulness metrics, using larger models such as RoBERTa decreases the disparity in the complexity of explanations, as it is most strongly visible when very few of the Grad, GxI, IG, and IGxI explanations show significant disparity in complexity on GECO. Nevertheless, despite this behavior, LIME and SHAP almost always result in disparity with RoBERTa on GECO, highlighting that the amount of disparity is not only a consequence of the model, but it varies with the explanation methods as well.


\subsubsection{Sensitivity Disparity}

Considering sensitivity, 85\% of runs on GECO results in disparities with considerable effect size, which is the highest among all evaluation metrics. Although not the highest among all metrics, 42\% of run on Stereotypes and 45\% on COMPAS also result in disparities with considerable effect size, indicating that such disparities persists across different datasets and explanation methods. 

Following the trend from the previous metrics, the larger RoBERTa model again results in the least amount of disparity in sensitivity on COMPAS and Stereotypes, highlighting again that larger models may be less, although not completely, prone to gender disparities in their explanations. However, this is not visible on the GECO datasets, where all runs result in statistically significant disparities in sensitivity. 

\begin{figure*}[t]
    \centering
    \begin{subfigure}{0.9\textwidth}
        \centering
        \includegraphics[width=\linewidth]{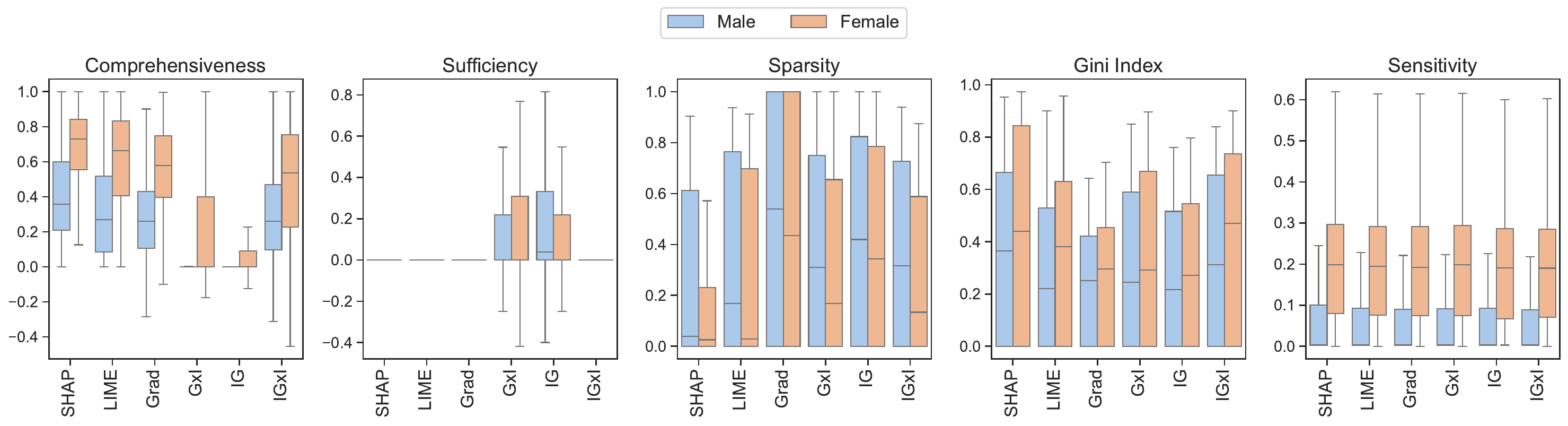}
        \caption{GECO-ALL}
        \label{fig:subfig1}
    \end{subfigure}

    \begin{subfigure}{0.9\textwidth}
        \centering
        \includegraphics[width=\linewidth]{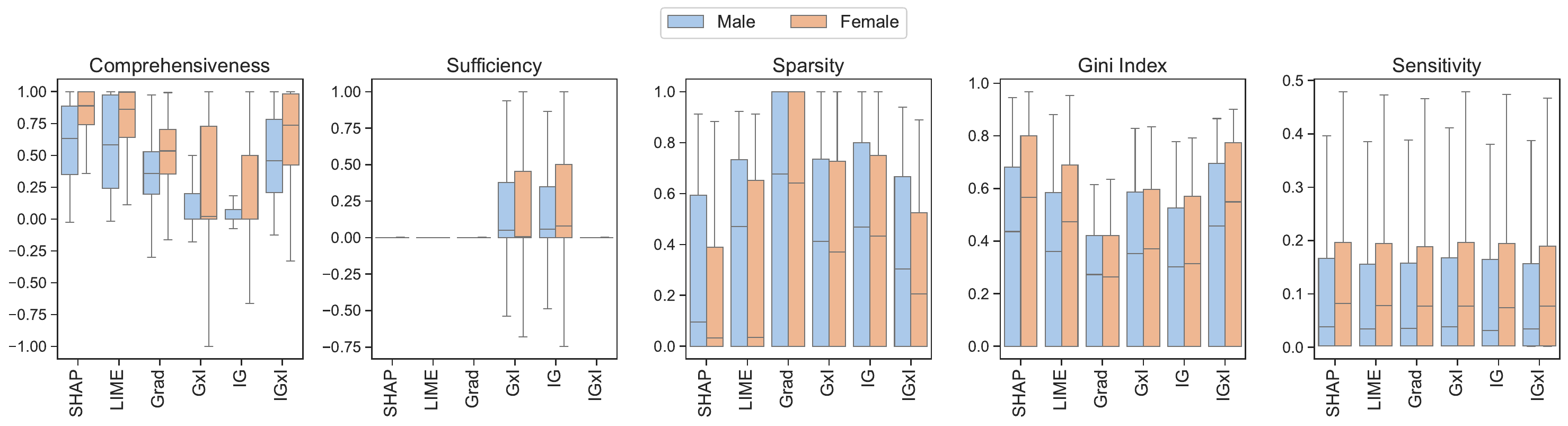}
        \caption{GECO-SUBJ}
        \label{fig:subfig1}
    \end{subfigure}
    
    \begin{subfigure}{0.9\textwidth}
        \centering
        \includegraphics[width=\linewidth]{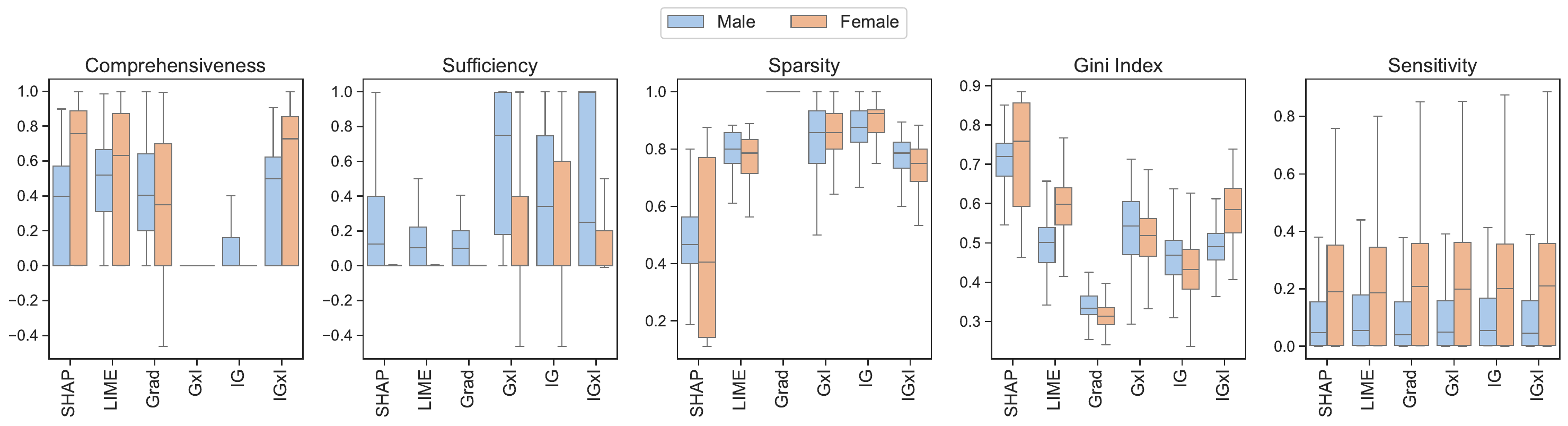}
        \caption{Stereotypes}
        \label{fig:subfig3}
    \end{subfigure}

    \begin{subfigure}{0.9\textwidth}
        \centering
        \includegraphics[width=\linewidth]{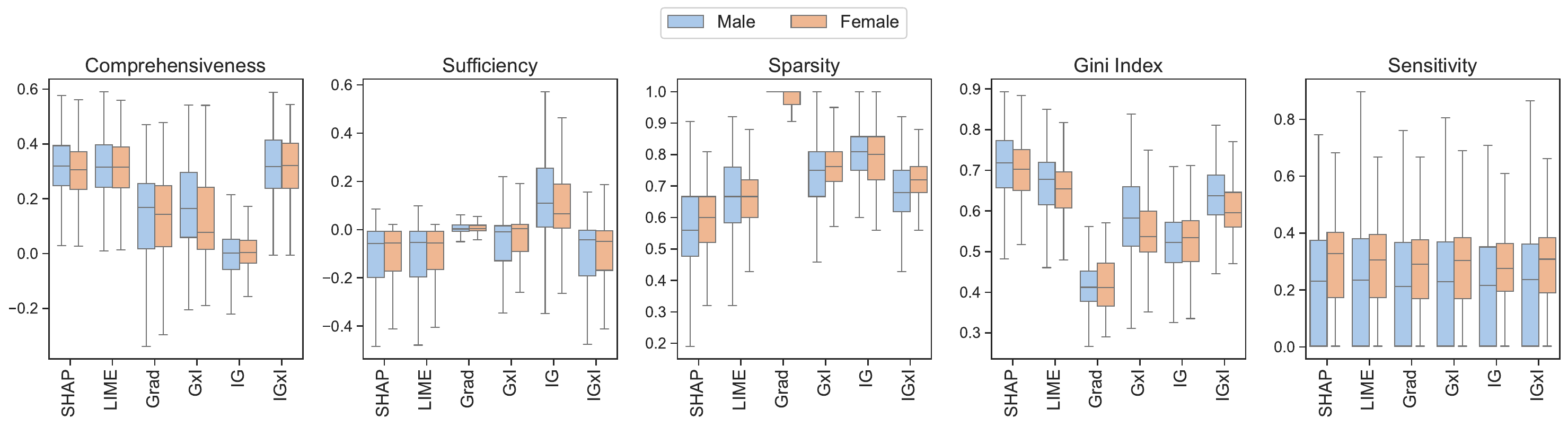}
        \caption{COMPAS}
        \label{fig:subfig4}
    \end{subfigure}
    \caption{\textbf{Box-plots of evaluation scores} obtained over 5 runs for each using TinyBERT on GECO, Stereotypes, and COMPAS, including the runs not resulting in statistically significant disparity.}
    \label{fig:boxplots}
\end{figure*}

{ 

\subsection{Disparities Across Datasets}

Overall, our results confirm the hypothesis that the GECO datasets would show the highest amount of disparity, since the sensitive attribute had a stronger influence on the task by way of being the label itself. As we decrease the impact of gender on the task, first with the Stereotypes and then with the COMPAS datasets, we observe less disparity but still a significant one. This indicates that the amount of disparity depends not only on the model or the explanation methods, but on the dataset as well. However, most crucially, this is not because the datasets are particularly biased but because the dataset determines the influence the sensitive attribute has on the predictions.

\subsection{Disparity when the Models are Trained from Scratch} \label{sec:train_scratch}

To eliminate the possibility that the disparity we observe is just a consequence of the data the models were pre-trained on, we now apply our pipeline to models trained only on the two variants of the GECO dataset. More specifically, using BERT and GPT-2, we initialize the models randomly and then train them either on GECO-ALL or GECO-SUBJ for 50 epochs. 

Table \ref{tab:random_init} displays the number of runs out of five resulting in statistically significant ($p \leq .05$) disparity, and highlights those that has a considerable effect size ($d \geq 0.2$) in bold. Similar to the results in Table in \ref{tab:new_all_table}, more than 80\% of runs for both models show significant gender disparity. Moreover, the direction of the disparity per metric, as indicated by the colors of cells in the table, also follows a similar pattern. For instance, in both sets of results, male sentences have explanations with higher soft sufficiency and sparsity scores, and female sentences have higher soft comprehensiveness scores. Thus, we conclude that even if trained only on an unbiased dataset such as GECO\footnote{{ We refer to GECO as "unbiased" as it does not distinguish between the two genders. Concretely, if the ground truth explanation words were masked, it would be impossible to determine which sentences were male sentences and which were female sentences. This is because with the masked inputs, the two sentences in each male-female pair appear identical, and there is no way to distinguish between the two genders since the dataset is perfectly balanced as well. This observation implies that there is no property of the dataset besides the gender words that affect the labels in any way. This is unlike a potentially biased dataset such as COMPAS where even if the gender of each data point was masked, the remaining features' statistics could be used to infer the masked genders to an extent.}}, the explanation methods frequently result in disparate treatments of the two genders. 
These results confirm that while datasets and models can influence the disparities in explanations, aligning with \citep{understading-disparity-2024}, 
they are not the sole cause and explanation methods themselves can contribute to these disparities.
}

\begin{table*}[t!]
\centering
\caption{\label{tab:random_init}\textbf{Training from scratch.} Counts of significant disparity with colors indicating direction of effect (red=female scores higher, blue=male scores higher)}
\small
\begin{tabular}{clrrrrrrr|rrrrrrr}
    \toprule
    &  & 
    \multicolumn{7}{c|}{GECO-ALL} & \multicolumn{7}{c}{GECO-SUBJ} \\
    \cmidrule{3-16}

      & &
    \multicolumn{4}{c}{Faithf.} & \multicolumn{2}{c}{Comp.} & \multicolumn{1}{c|}{Rbst.} &
     \multicolumn{4}{c}{Faithf.} & \multicolumn{2}{c}{Comp.} & \multicolumn{1}{c}{Rbst.} \\
    \cmidrule{3-16}
    
    \textbf{Model} & \textbf{Method} & \rot{90}{Compr.} & \rot{90}{Suff.} & \rot{90}{Soft Compr.} & \rot{90}{Soft Suff.} & \rot{90}{Gini} & \rot{90}{Spars.} & \rot{90}{Sens.} & \rot{90}{Compr.} & \rot{90}{Suff.} & \rot{90}{Soft Compr.} & \rot{90}{Soft Suff.} & \rot{90}{Gini} & \rot{90}{Spars.} & \rot{90}{Sens.} \\
    \midrule
    &  Grad &  \textbf{\cellcolor{red!50.0}5} & {\cellcolor{red!40.0}4} & \textbf{\cellcolor{red!40.0}4} & \textbf{\cellcolor{red!40.0}4} & \textbf{\cellcolor{blue!40.0}4} & \textbf{\cellcolor{blue!20.0}2} & \textbf{\cellcolor{blue!40.0}4} & \textbf{\cellcolor{red!50.0}5} & {\cellcolor{blue!40.0}4} & \textbf{\cellcolor{red!40.0}4} & \textbf{\cellcolor{red!50.0}5} & {\cellcolor{red!30.0}3} & {\cellcolor{blue!20.0}2} & \textbf{\cellcolor{blue!20.0}2} \\
    &  GxI &  \textbf{\cellcolor{red!50.0}5} & \textbf{\cellcolor{blue!40.0}4} & \textbf{\cellcolor{red!40.0}4} & \textbf{\cellcolor{red!50.0}5} & {\cellcolor{blue!40.0}4} & {\cellcolor{red!50.0}5} & \textbf{\cellcolor{blue!40.0}4} & \textbf{\cellcolor{red!50.0}5} & \textbf{\cellcolor{blue!50.0}5} & \textbf{\cellcolor{red!30.0}3} & \textbf{\cellcolor{red!40.0}4} & {\cellcolor{red!40.0}4} & {\cellcolor{red!40.0}4} & \textbf{\cellcolor{blue!20.0}2} \\
    &  IG &  \textbf{\cellcolor{red!30.0}3} & \textbf{\cellcolor{blue!30.0}3} & \textbf{\cellcolor{red!50.0}5} & \textbf{\cellcolor{red!50.0}5} & \textbf{\cellcolor{red!40.0}4} & \textbf{\cellcolor{red!30.0}3} & \textbf{\cellcolor{blue!40.0}4} & \textbf{\cellcolor{red!30.0}3} & \textbf{\cellcolor{blue!40.0}4} & \textbf{\cellcolor{red!50.0}5} & \textbf{\cellcolor{red!30.0}3} & \textbf{\cellcolor{red!50.0}5} & \textbf{\cellcolor{red!50.0}5} & \textbf{\cellcolor{blue!20.0}2} \\
    \multirow[t]{6}{*}{\rotatebox[origin=c]{45}{\textbf{BERT}}}  &  IGxI &  \textbf{\cellcolor{red!40.0}4} & {\cellcolor{red!40.0}4} & \textbf{\cellcolor{red!50.0}5} & \textbf{\cellcolor{red!50.0}5} & \textbf{\cellcolor{red!40.0}4} & \textbf{\cellcolor{red!40.0}4} & \textbf{\cellcolor{blue!40.0}4} & \textbf{\cellcolor{red!50.0}5} & \textbf{\cellcolor{red!50.0}5} & \textbf{\cellcolor{red!30.0}3} & \textbf{\cellcolor{red!30.0}3} & \textbf{\cellcolor{red!40.0}4} & \textbf{\cellcolor{red!40.0}4} & \textbf{\cellcolor{blue!30.0}3} \\
    &  LIME &  \textbf{\cellcolor{red!40.0}4} & \textbf{\cellcolor{blue!40.0}4} & \textbf{\cellcolor{red!50.0}5} & \textbf{\cellcolor{red!50.0}5} & \textbf{\cellcolor{red!30.0}3} & {\cellcolor{red!40.0}4} & \textbf{\cellcolor{blue!40.0}4} & \textbf{\cellcolor{red!50.0}5} & {\cellcolor{blue!50.0}5} & \textbf{\cellcolor{red!30.0}3} & \textbf{\cellcolor{red!30.0}3} & \textbf{\cellcolor{red!50.0}5} & \textbf{\cellcolor{red!50.0}5} & \textbf{\cellcolor{blue!30.0}3} \\
    &  SHAP &  \textbf{\cellcolor{red!50.0}5} & \textbf{\cellcolor{blue!40.0}4} & \textbf{\cellcolor{red!50.0}5} & \textbf{\cellcolor{red!50.0}5} & {\cellcolor{blue!50.0}5} & {\cellcolor{red!50.0}5} & \textbf{\cellcolor{blue!40.0}4} & \textbf{\cellcolor{red!40.0}4} & {\cellcolor{blue!50.0}5} & \textbf{\cellcolor{red!50.0}5} & \textbf{\cellcolor{red!50.0}5} & \textbf{\cellcolor{red!50.0}5} & \textbf{\cellcolor{red!40.0}4} & \textbf{\cellcolor{blue!30.0}3} \\
    \cmidrule{1-16}
     &  Grad &  \textbf{\cellcolor{red!50.0}5} & \textbf{\cellcolor{blue!40.0}4} & \textbf{\cellcolor{red!50.0}5} & \textbf{\cellcolor{red!50.0}5} & {\cellcolor{red!20.0}2} & \textbf{\cellcolor{red!20.0}2} & {\cellcolor{red!40.0}4} & \textbf{\cellcolor{red!50.0}5} & \textbf{\cellcolor{blue!30.0}3} & \textbf{\cellcolor{red!50.0}5} & \textbf{\cellcolor{red!50.0}5} & \textbf{\cellcolor{blue!40.0}4} & \textbf{\cellcolor{blue!40.0}4} & \textbf{\cellcolor{red!10.0}1} \\
    &  GxI &  \textbf{\cellcolor{red!40.0}4} & \textbf{\cellcolor{blue!50.0}5} & \textbf{\cellcolor{red!50.0}5} & \textbf{\cellcolor{red!50.0}5} & \textbf{\cellcolor{red!50.0}5} & \textbf{\cellcolor{red!40.0}4} & {\cellcolor{red!30.0}3} & \textbf{\cellcolor{red!50.0}5} & \textbf{\cellcolor{blue!40.0}4} & \textbf{\cellcolor{red!50.0}5} & \textbf{\cellcolor{red!50.0}5} & \textbf{\cellcolor{blue!30.0}3} & {\cellcolor{blue!40.0}4} & \textbf{\cellcolor{red!10.0}1} \\
    &  IG &  \textbf{\cellcolor{red!50.0}5} & \textbf{\cellcolor{blue!40.0}4} & \textbf{\cellcolor{red!50.0}5} & \textbf{\cellcolor{red!50.0}5} & \textbf{\cellcolor{red!50.0}5} & \textbf{\cellcolor{red!50.0}5} & {\cellcolor{blue!20.0}2} & \textbf{\cellcolor{red!50.0}5} & \textbf{\cellcolor{blue!50.0}5} & \textbf{\cellcolor{red!50.0}5} & \textbf{\cellcolor{red!50.0}5} & \textbf{\cellcolor{red!50.0}5} & \textbf{\cellcolor{red!30.0}3} & \textbf{\cellcolor{red!20.0}2} \\
    \multirow[t]{6}{*}{\rotatebox[origin=c]{45}{\textbf{GPT2}}} &  IGxI &  \textbf{\cellcolor{red!50.0}5} & \textbf{\cellcolor{blue!30.0}3} & \textbf{\cellcolor{red!50.0}5} & \textbf{\cellcolor{red!50.0}5} & \textbf{\cellcolor{red!50.0}5} & \textbf{\cellcolor{red!50.0}5} & {\cellcolor{red!30.0}3} & \textbf{\cellcolor{red!50.0}5} & \textbf{\cellcolor{blue!50.0}5} & \textbf{\cellcolor{red!50.0}5} & \textbf{\cellcolor{red!50.0}5} & \textbf{\cellcolor{red!40.0}4} & \textbf{\cellcolor{red!40.0}4} & \textbf{\cellcolor{red!20.0}2} \\
    &  LIME &  \textbf{\cellcolor{red!50.0}5} & \textbf{\cellcolor{blue!50.0}5} & \textbf{\cellcolor{red!50.0}5} & \textbf{\cellcolor{red!50.0}5} & \textbf{\cellcolor{red!40.0}4} & \textbf{\cellcolor{red!30.0}3} & {\cellcolor{blue!20.0}2} & \textbf{\cellcolor{red!50.0}5} & {\cellcolor{blue!50.0}5} & \textbf{\cellcolor{red!50.0}5} & \textbf{\cellcolor{red!50.0}5} & {\cellcolor{red!40.0}4} & {\cellcolor{red!30.0}3} & \textbf{\cellcolor{red!20.0}2} \\
    &  SHAP &  \textbf{\cellcolor{red!50.0}5} & \textbf{\cellcolor{blue!50.0}5} & \textbf{\cellcolor{red!50.0}5} & \textbf{\cellcolor{red!50.0}5} & \textbf{\cellcolor{red!50.0}5} & \textbf{\cellcolor{red!40.0}4} & {\cellcolor{red!30.0}3} & \textbf{\cellcolor{red!50.0}5} & {\cellcolor{red!50.0}5} & \textbf{\cellcolor{red!50.0}5} & \textbf{\cellcolor{red!50.0}5} & \textbf{\cellcolor{red!40.0}4} & \textbf{\cellcolor{red!30.0}3} & {\cellcolor{red!20.0}2} \\
    \bottomrule
\end{tabular}
\end{table*}

\subsection{Implications and Considerations for Researchers and Practitioners}

Although disparity results can vary between metrics, all explanation methods under study consistently exhibit significant explanation disparities with considerable effect sizes across all included metrics. These results underscore the need for stakeholders\footnote{According to the Merriam-Webster dictionary, a stakeholder is defined, among other things, as someone “who is involved in or affected by a course of action.” \citep{dictionary}, so following \citep{langer-xai-stakeholder-perspective-2021}, we use this a broad term but specify some particular stakeholder categories when necessary.} (e.g., practitioners, developers, researchers)
to consider general and metric-specific explanation disparities when using explanations for PLMs outputs to make informed decisions, depending on their use case. 
{ 

\textbf{Practitioners} use explainability methods to interpret a model's decision-making process, debug the model, or improve its performance. As previously discussed, the convenience and ease of use provided by explainability frameworks make them popular among developers seeking explanations for real-world applications. However, directly applying post-hoc methods or relying on frameworks that support them can mislead developers when evaluating the model, particularly in gender-related tasks, leading to biased decisions and critical outcomes for both the system under development and any subsequent projects that utilize such frameworks.  
Practitioners should therefore recognize that these methods can exhibit significant gender disparities. We recommend they carefully consider such disparities based on the explanation properties most relevant to their use case. As discussed earlier, each explanation property has different implications. Accordingly, practitioners must assess which properties are most critical in their specific contexts. 
For example, certain disparities in explanations can be more critical for some stakeholder groups than for others. For instance, explanations with complexity disparity might not be critical for developers. However, it can be very relevant for laypeople who often need simple and easily understandable explanations. On the other side, significant disparity in faithfulness represents a major concern for all stakeholder groups as it implies explanations that inaccurately reflect the model's decision-making process between subgroups, which could result in critical consequences, similar to the example presented in section \ref{sec:intro}.
We urge practitioners to thoroughly audit the properties of explanations for each subgroup.
}

\textbf{Researchers} can benefit from our open-source pipeline to develop new explanation methods and evaluation metrics and identify the reasons behind the disparity we observe. Practitioners can also use this pipeline to run tests to identify potential failure modes of the methods they are using. For example, detecting that explanation quality varies significantly by gender in a task where it should be gender-neutral raises a potential red flag, especially in gender-sensitive contexts. {  Thus, we call upon researchers and developers to account for gender disparities in post-hoc explanations when introducing new libraries and frameworks that employ these methods.}

{ 
For \textbf{developers} of AI systems, particularly those integrating PLMs, employing explainability methods that exhibit gender disparities in systems can lead to non-compliance with transparency requirements outlined in regulations such as the EU AI Act \citep{eu-ai-act}. This risk is particularly pronounced in high-risk settings, where biased explanations can undermine the system's fairness, disadvantage certain subgroups, and impose significant liability on developers and deployers of such systems. 

As post-hoc methods are widely used across various AI applications, \textbf{end-users}
, such as doctors, who receive explanations generated by these methods, should be aware that they may exhibit gender disparities, particularly in gender-sensitive tasks or use cases.

For \textbf{regulators and policymakers}, our findings emphasize the importance of explicitly integrating explanation fairness as a requirement in both existing and future regulations alongside model and data fairness. 
}




\section{Conclusion}

In this paper, we presented the first study investigating disparities in the quality of post-hoc feature attribution methods for language models across subgroups, focusing specifically on gender as a protected attribute. We showed that every investigated explanation method presents a significant degree of bias across various metrics, even when the models are trained from scratch on an unbiased dataset, with the most pronounced disparities emerging in faithfulness and sensitivity. These results underscore the importance of going beyond model-level fairness and scrutinizing the fairness of explanations themselves.

Despite the limitations discussed in Section \ref{sec:limitations}, this work can serve as an essential foundation for researchers and practitioners seeking to evaluate existing and novel methods of interpreting language models. By unveiling potential gaps in how explanation quality varies for different demographic groups and metrics, we highlight the broader need for fairness-promoting algorithms that address explanation-level bias. Building on recent efforts to mitigate model bias \citep{fairness-xai-interplay-2024}, we advocate for fairness-focused strategies aimed at reducing disparities in explanation performance.

Looking forward, there are multiple avenues for \textbf{future work}:
\begin{itemize}
    \item Extending methods and metrics: Incorporate new approaches \citep{slalom2025, atman2023} or additional implementations of existing metrics.
     \item Broadening data coverage: 
     Generate synthetic datasets or augment existing textual datasets to capture protected attributes beyond gender, ensuring compatibility with our disparity measurement pipeline. Additionally, we plan to expand our dataset collection by integrating further datasets addressing gender bias in NLP tasks, such as WinoBias \cite{zhao-etal-2018-WinoBias}, while still being aware of the shortcomings of such datasets \cite{blodgett-etal-2021-stereotyping-dataset-pifall}.
    \item Combining quantitative with human-based evaluation: Complement the standard metrics with human-grounded assessments to evaluate explanations \citep{co12-xai-properties-2023,longo-manifesto-2024} to capture nuanced disparities early.
\end{itemize}
Our findings point to the need for deeper theoretical and empirical investigation into the causes of explanation bias and the contribution of each cause, whether stemming from the explanation methods themselves, the fine-tuning process, or dataset design. A better understanding of these underlying mechanisms will be pivotal for developing robust mitigation strategies and ensuring that explanation fairness is upheld alongside transparency and predictive fairness.

\section{Limitations} \label{sec:limitations} 
The main limitation of our evaluation is that it is limited to gender disparity and binary classification tasks. More insights into the disparities amplified by the explanation methods can be gained by analyzing different sensitive attributes such as race, as well as other tasks, such as text generation. Our evaluations are also limited to transformer-based models, although such models currently see the highest use. 
 We also acknowledge that the design process for the synthetic Stereotypes dataset could benefit from recommendations in the literature to avoid potential pitfalls associated with evaluation corpus design \cite{blodgett-etal-2020-language-dataset-pitfall, blodgett-etal-2021-stereotyping-dataset-pifall}.
Finally, evaluating explanations with respect to desirable properties such as faithfulness and robustness is an active research area itself, with new evaluation methods frequently being proposed to address the shortcomings of existing methods \citep{jacovi-goldberg-2020-faithfulness,lyu-2024-towards-faithful-model-explanation-NLP, hsia-2024-goodharts-nlp-explanations}. Thus, our analysis is also limited by the current state of the evaluation literature and would benefit from future developments.


\begin{acks}
We would like to thank the anonymous reviewers for their helpful suggestions. This research has been supported by the German Federal Ministry of Education and Research (BMBF) grant 01IS23069 Software Campus 3.0 (TU München). NF is supported by grant 16KIS2047 "VERANDA" (TU Berlin), and BIFOLD 24B.
\end{acks}

\bibliographystyle{ACM-Reference-Format}
\bibliography{custom}


\appendix

\section{Definition and Formulation of Evaluation Metrics} \label{app:metric-defs}

\subsection{Comprehensiveness} 

Measures how relevant the tokens assigned high-importance are for classification. Let $f_j$ be the output probability for the correct class $j$. Top-$k$ tokens $r$ are removed and the difference $f_j(x) - f_j(x \setminus r)$ is the comprehensiveness value. In ferret, comprehensiveness is measured using the area over perturbation curve (AOPC) that is computed by varying $k$ (the number of tokens to remove) by varying the threshold such that the tokens with an importance score above the threshold are removed, and averaging the resulting comprehensiveness values. The resulting values thus lie in the interval [0,1]. We vary threshold from 0.1 to 1.0 in increments of 0.1. A high value indicates a significant change in the model's output, which implies that the removed tokens were important for classification. Then we conclude that an explanation successfully captures the relevant tokens if it has a high comprehensiveness value.

\subsection{Sufficiency}
As opposed to comprehensiveness, only the top-$k$ tokens $r$ are input to the model and the sufficiency value is the difference $f_j(x) - f_j(r)$ in the model's output. A small value indicates that only the tokens assigned high importance were enough to obtain the same output, and hence that the explanation was able to capture the most relevant tokens. Then the number $k$ is varied similar to the comprehensiveness metric, except this time removing the tokens with importance scores below the threshold, and the AOPC is computed by averaging the resulting sufficiency values, with the final values between 0 and 1. 

\subsection{Soft Sufficiency and Comprehensiveness}

Removing tokens entirely can lead to out-of-distribution inputs, meaning that the explanations are evaluated on kinds of inputs the model never saw during training and is unlikely to see in real use. To reduce this difference between the actual inputs and those used in evaluation, \citep{aopc-limitations-2023} instead propose to mask a fraction of each token's embeddings based on that token's importance score. For a the vector representation $\mathbf{x}$ of a token with importance score $s$ normalized between 0 and 1, the input is perturbed to obtain $\mathbf{x}'$ such that
\begin{equation}
    \mathbf{x}' = \mathbf{x} \odot \mathbf{e}, 
    \quad \mathbf{e}_i \sim \text{Ber}(q)
\end{equation}
with $q=s$ if the elements are to be retained (for sufficiency) and $q = 1-s$ if they are to be removed (for comprehensiveness). Finally for original and perturbed sentences $\mathbf{X}$ and $\mathbf{X}'$ with true class $y$, soft sufficiency and comprehensiveness are defined as
\begin{align}
    \text{Soft-S} &= 1 - \max(0, p(y \vert X) - p(y \vert \mathbf{X}')) \in [0, 1] \\
    \text{Soft-C} &= \max(0, p(y \vert X) - p(y \vert \mathbf{X}')) \in [0, 1]
\end{align}
where $p$ denotes the model output logits.

\subsection{Sparsity}

For a given explanation vector $(s_1,...,s_n)$, we compute the share of scores exceeding a threshold $\tau$ (0.1 for our experiments) in absolute value:
\begin{equation}
    \text{Sparsity} := \frac{1}{n}\sum_{i=1}^n \mathbf{1}\left[ \left\vert s_i \right\vert \geq \tau \right]
    \in [0,1]
\end{equation}
where $\mathbf{1}$ denotes the indicator function. {  Lower non-zero values are preferred as they indicate only a few tokens were assigned high scores, which makes the explanation easier to understand. Sparsity of zero is not desired since it means all tokens were assigned relatively low scores with respect to the threshold.}

\subsection{Gini Index}

For the explanation vector $\mathbf{s} = (s_1,...,s_n)$ sorted in an ascending way with respect to the scores' absolute values and $\mathbf{k} = (k_1, ..., k_n)$ denoting the indices of the original elements in the sorted vector, we compute 
\begin{equation}
    \text{Gini Index} := 1 - 2 \sum_{i=1}^n \frac{s_i}{\Vert \mathbf{s} \Vert_1} \cdot \frac{n - k_i + 0.5}{n}
    \in [0,1]
\end{equation}
with higher values (more sparse) being preferred. 



\subsection{Sensitivity}

Given input $x$ and model $f$ with explainer $\Phi$, we find the input $y$ within a ball of radius $r$ around $x$ such that the change in the explanation is maximized (i.e. the worst-case perturbation). A lower worst-case difference indicates the explanation method is more robust to small perturbations:
\begin{equation}
\text{Sensitivity} = \max_{y; \Vert x - y \Vert \leq r}
\frac{\Vert \Phi(f,y) - \Phi(f, x) \Vert}{\Vert \Phi(f, x) \Vert} 
\in [0, \infty).
\end{equation}
We use a projected gradient descent (PGD) \citep{madry2017towards} attack in which the input is perturbed in the direction of the gradient maximizing the prediction error, and projected back onto the ball after each gradient step.

\section{Reproducibility}

{  Our end-to-end pipeline is designed to be easily reproducible. We base our experiments on the publicly available GECO \citep{wilming2024gecobench} and COMPAS datasets \citep{angwin2016compas}, as well as the synthetic Stereotypes dataset we create and make public.} We use the publicly available models from Huggingface (see Table \ref{tab:models}) running on a single NVIDIA V100 GPU. Including fine-tuning, and generating and evaluating explanations, one model/dataset run takes between 40-60 minutes without the sensitivity metric and ~10 hours with sensitivity. We use the open source ferret library \citep{attanasio-etal-2023-ferret} for implementations of the explanation methods and metrics. For the metrics not available in ferret, we either use and include in our codebase other publicly available implementations with appropriate licenses, or provide our own implementations. 

\begin{table}[h!]
    \centering
    \caption{\label{tab:models}Information about the models used in our experiments. The names are hyperlinks directing to their respective HuggingFace Hub pages.}
    \begin{tabular}{lll}
        \toprule
         \textbf{Name} & \textbf{Type} & \textbf{Num. Params} \\
         \midrule
         \href{https://huggingface.co/huawei-noah/TinyBERT_General_4L_312D}{TinyBERT} & Encoder-only  & 14,350,874  \\
         \href{https://huggingface.co/facebook/FairBERTa}{FairBERTa} & Encoder-only & 124,647,170\\
         \href{https://huggingface.co/nlptown/bert-base-multilingual-uncased-sentiment}{BERT} & Encoder-only & 167,357,954 \\
         \href{https://huggingface.co/openai-community/gpt2}{GPT-2} & Decoder-only & 124,442,112 \\
         \href{https://huggingface.co/FacebookAI/roberta-large}{RoBERTa-large} & Encoder-only & 355,359,744 \\
         \bottomrule
    \end{tabular}
\end{table}

{ 
\section{Prompting Claude for the Stereotypes Dataset} \label{app:prompts}

To create our Stereotypes dataset, we prompt Claude 3.5 Sonnet \citep{anthropic2024claude}, the most recent version as of November 2024. Rather than using a single prompt, we start with an initial prompt, and then iterate in a few steps of conversation depending on the quality of the sentences generated. Once the sentences fulfill our requirements, we repetitively ask Claude to generate a number of sentences. 

The initial prompt is:
\begin{displayquote}
\textit{I want you to generate a small dataset. It will consist of pairs of sentences. The only difference between the sentences in each pair will be the subject's gender. E.g.:
\begin{itemize}
    \item He is a doctor.
    \item She is a doctor.
\end{itemize}
The second characteristic of the dataset is that the first sentence in each pair will express a stereotype towards one gender. So the second sentence will be the same, just with the gender flipped, and the stereotype naturally won't hold for that gender. E.g.
\begin{itemize}
    \item She was a bad doctor, no surprises.
    \item He was a bad doctor, no surprises.
\end{itemize}
Do you understand? Generate one sentence pair so I can see if you get the task.}
\end{displayquote}
After this prompt, we give feedback for two steps until the outputs are of desired quality. Our feedback consists of the instructions
\begin{displayquote}
    \textit{You get the point but the examples you generated are not very good. Generate a few more and I will tell you the  best. Then we will refine.}
\end{displayquote}
and
\begin{displayquote}
    \textit{But the stereotypes are not explicitly obvious in the sentences. I want them to be more clear. Something like "I was surprised to see a woman doctor articulate herself so well."}
\end{displayquote}
}

\section{Definition of Cohen's $d$} \label{app:cohen}

To quantify the effect size in our experiments we use the Cohen's $d$ metric defined as 
\begin{equation}
    d = \frac{\bar x_M - \bar x_F}{s}
    \quad \text{with} \quad
    s = \sqrt{\frac{\sigma_M^2 + \sigma_F^2}{2}}
\end{equation}
with $\bar x_M, \bar x_F$ average male/female scores and $\sigma_M^2, \sigma_F^2$ the variances.

\section{Bias Analysis} \label{sec:bias-analysis}

To ensure that the bias we observe is independent of the model, we quantify the gender bias in each of our models through a bias analysis after fine-tuning, with the results displayed below. The true positive rate (TPR), true negative rate (TNR), and the average prediction difference (APD) \citep{jentzsch2022gender} defined as follows:

\begin{align}
    \text{TPR} &= \frac{\text{Accurately predicted male}}{\text{Total male}} \\
    \text{TNR} &= \frac{\text{Accurately predicted female}}{\text{Total female}} \\
    \text{APD} &= \mathbb{E}_{x \sim \mathcal{D}} \left\vert 
            f_{\text{sm}}^M\left(x_M\right) - f_{\text{sm}}^F\left(x_F\right)   
    \right\vert
\end{align}
where $x_M$ and $x_F$ are the male and female versions of the same input, and $\mathcal{D}$ denotes the dataset, and $f_{\text{sm}}$ the softmax output probabilities of the model for the male ($M$) and female ($F$) classes.

Thus a discrepancy between the TPR and TNR values indicate the model is better at identifying one gender than the other, and while a high APD value can indicate a discrepancy between the subgroup accuracies, it might also indicate that the model is more certain (i.e. higher softmax probabilities
when making predictions for one gender compared to the other one.

Table \ref{tab:bias_analysis} displays the bias analysis results for our four models averaged over 5 runs. The average predictions differences are smaller for the ALL dataset compared to the slightly harder SUBJ dataset. Nevertheless, all models achieve almost-perfect accuracy on ALL and very high accuracy on SUBJ. While accuracies for male sentences is slightly higher for all models as the TPR values are higher than TNR values, the very small differences often within $\pm$ one standard deviation and in in the order of $0.01$ leads us to believe there is no significant bias inherent in the models after fine-tuning. 

\begin{table}[h!]
\centering
\definecolor{lightgray}{rgb}{0.5,0.5,0.5}
\caption{\label{tab:bias_analysis}Bias analysis results after fine-tuning each model, averaged over 5 runs (TPR: true positive rate, TNR, true negative rate, APD: average prediction difference).}
    \begin{tabular}{lllll}
    \toprule
     \textbf{Dataset} & \textbf{Model} & \textbf{TPR} & \textbf{TNR} & \textbf{APD} \\
    \midrule
    \multirow[t]{5}{*}{ALL} & BERT & 0.996\textcolor{lightgray}{$_{0.001}$} & 0.991\textcolor{lightgray}{$_{0.001}$} & 0.006\textcolor{lightgray}{$_{0.000}$} \\
     & FairBERTa & 0.997\textcolor{lightgray}{$_{0.000}$} & 0.996\textcolor{lightgray}{$_{0.001}$} & 0.002\textcolor{lightgray}{$_{0.004}$} \\
     & GPT2 & 0.997\textcolor{lightgray}{$_{0.000}$} & 0.994\textcolor{lightgray}{$_{0.000}$} & 0.004\textcolor{lightgray}{$_{0.000}$} \\
     & TinyBERT & 0.992\textcolor{lightgray}{$_{0.003}$} & 0.988\textcolor{lightgray}{$_{0.001}$} & 0.013\textcolor{lightgray}{$_{0.001}$} \\
     & RoBERTa & 0.979\textcolor{lightgray}{$_{0.020}$} & 0.975\textcolor{lightgray}{$_{0.036}$} & 0.054\textcolor{lightgray}{$_{0.031}$} \\
    \midrule
    \multirow[t]{5}{*}{SUBJ} & BERT & 0.985\textcolor{lightgray}{$_{0.001}$} & 0.979\textcolor{lightgray}{$_{0.001}$} & 0.029\textcolor{lightgray}{$_{0.002}$} \\
     & FairBERTa & 0.983\textcolor{lightgray}{$_{0.003}$} & 0.960\textcolor{lightgray}{$_{0.008}$} & 0.050\textcolor{lightgray}{$_{0.005}$} \\
     & GPT2 & 0.979\textcolor{lightgray}{$_{0.003}$} & 0.967\textcolor{lightgray}{$_{0.003}$} & 0.046\textcolor{lightgray}{$_{0.005}$} \\
     & TinyBERT & 0.984\textcolor{lightgray}{$_{0.001}$} & 0.971\textcolor{lightgray}{$_{0.004}$} & 0.039\textcolor{lightgray}{$_{0.001}$} \\
     & RoBERTa & 0.895\textcolor{lightgray}{$_{0.076}$} & 0.973\textcolor{lightgray}{$_{0.009}$} & 0.170\textcolor{lightgray}{$_{0.098}$} \\
    \midrule
    \multirow[t]{5}{*}{COMPAS} & BERT & 0.626\textcolor{lightgray}{$_{0.000}$} & 0.720\textcolor{lightgray}{$_{0.000}$} & 0.230\textcolor{lightgray}{$_{0.000}$} \\
     & FairBERTa & 0.654\textcolor{lightgray}{$_{0.000}$} & 0.720\textcolor{lightgray}{$_{0.000}$} & 0.237\textcolor{lightgray}{$_{0.000}$} \\
     & GPT2 & 0.590\textcolor{lightgray}{$_{0.000}$} & 0.735\textcolor{lightgray}{$_{0.000}$} & 0.249\textcolor{lightgray}{$_{0.000}$} \\
     & TinyBERT & 0.595\textcolor{lightgray}{$_{0.035}$} & 0.749\textcolor{lightgray}{$_{0.021}$} & 0.244\textcolor{lightgray}{$_{0.017}$} \\
     & RoBERTa & 0.000\textcolor{lightgray}{$_{0.000}$} & 1.000\textcolor{lightgray}{$_{0.000}$} & 0.115\textcolor{lightgray}{$_{0.002}$} \\
    \midrule
    \multirow[t]{5}{*}{Stereotypes} & BERT & 0.994\textcolor{lightgray}{$_{0.000}$} & 0.997\textcolor{lightgray}{$_{0.000}$} & 0.003\textcolor{lightgray}{$_{0.000}$} \\
     & FairBERTa & 0.994\textcolor{lightgray}{$_{0.000}$} & 1.000\textcolor{lightgray}{$_{0.000}$} & 0.006\textcolor{lightgray}{$_{0.000}$} \\
     & GPT2 & 0.997\textcolor{lightgray}{$_{0.000}$} & 1.000\textcolor{lightgray}{$_{0.000}$} & 0.002\textcolor{lightgray}{$_{0.000}$} \\
     & TinyBERT & 1.000\textcolor{lightgray}{$_{0.000}$} & 1.000\textcolor{lightgray}{$_{0.000}$} & 0.000\textcolor{lightgray}{$_{0.000}$} \\
     & RoBERTa & 1.000\textcolor{lightgray}{$_{0.000}$} & 0.800\textcolor{lightgray}{$_{0.400}$} & 0.001\textcolor{lightgray}{$_{0.003}$} \\
    \bottomrule
    \end{tabular}
\end{table}


\section{Experimental Pipeline} \label{sec:experimental-pipeline}

To obtain our results, for each of our models and datasets, we split the dataset into an 80/20 train/test split with balanced classes. We download pre-trained model weights \footnote{As discussed in section \ref{sec:train_scratch}, for the experiments where we train models from scratch, we initialize the models randomly and then train them either on GECO-ALL or GECO-SUBJ for 50  epochs (1 epoch for RoBERTa to ensure high test accuracy without overfitting).} from Huggingface, and update all weights during training for 50 epochs for the GECO dataset, 5 for COMPAS, and 10 epochs for the Stereotypes dataset. We use the AdamW optimizer \citep{loshchilov2017decoupled} with initial learning rate 0.001 and a linear learning rate schedule with 500 warm-up steps. Since our tasks are binary classification tasks, we use the binary cross-entropy loss. We observed that after one epoch of fine-tuning, the models perform hardly better than random guessing, so we trained each model for a larger number of epochs to achieve a high test accuracy without overfitting. We repeat this process 5 times for each model and dataset pair to report aggregate results. 

\begin{algorithm}[htbp]

\caption{Experimental Pipeline}\label{alg:exp}
\KwData{\\ Dataset $(D_F, D_M)$ w/ female/male subsets  \\
        Fine-tuned models $f_1,...,f_4$ \\
        Explanation methods $e_1,...,e_6$ \\
        Evaluation metrics $m_1,...,m_6$}
Let $f,e,m$ denote an arbitrary model, explainer, and metric. \\
Init lists of male/female scores $S_M, S_F$. \\
\For{$(x_i^M, x_i^F)$ in $(D_M, D_F)$}{
  $s_M \gets (m \circ e \circ f)(x_i^M)$ \\
  $s_F \gets (m \circ e \circ f)(x_i^F)$ \\
  $S_M$.\texttt{append}($s_M$) \\
  $S_F$.\texttt{append}($s_F$) \\
}
$p \gets \texttt{Mann-Whitney-U}(S_M, S_F)$ \\
\If{$p\leq .05$}{
    $d \gets (\bar S_M - \bar S_F)\left( \sqrt{\frac{\sigma_M^2 + \sigma_F ^ 2}{2}} \right)^{-1}$ \\
    \textbf{Return} "significant" with effect size $d$. \\
} \Else{
    \textbf{Return} "not significant".
}
\end{algorithm}

\section{Additional Results} \label{app:additional-results}

\begin{figure*}[t]
    \centering
    \begin{subfigure}{0.45\textwidth}
        \centering
        \includegraphics[width=\linewidth]{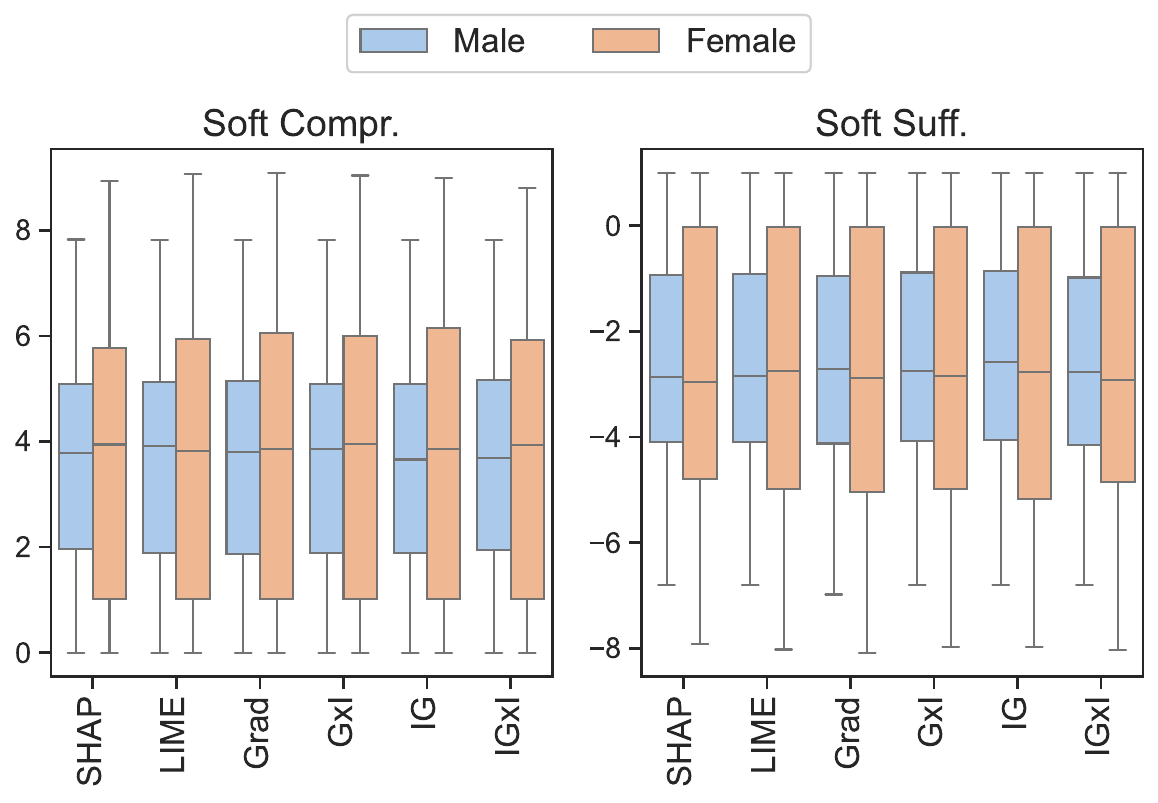}
        \caption{GECO-ALL}
        \label{fig:soft_subfig1}
    \end{subfigure}
    \begin{subfigure}{0.45\textwidth}
        \centering
        \includegraphics[width=\linewidth]{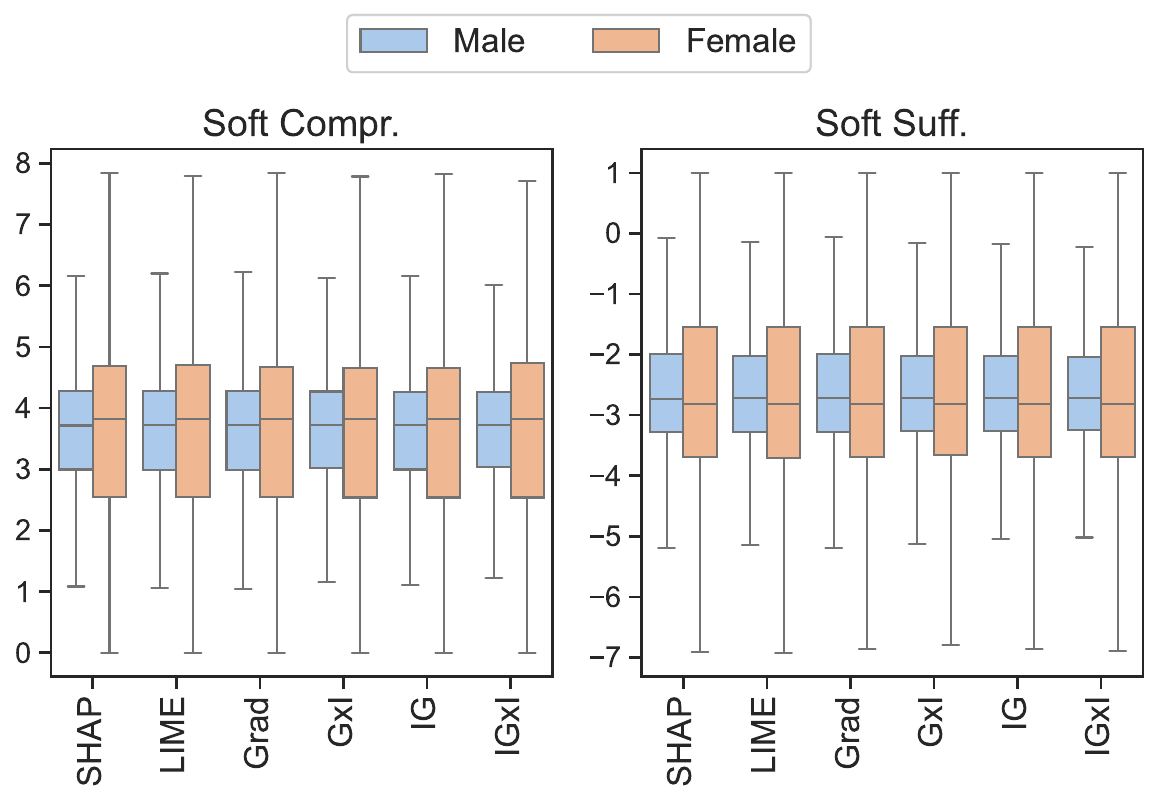}
        \caption{GECO-SUBJ}
        \label{fig:soft_subfig3}
    \end{subfigure}
    
    \begin{subfigure}{0.45\textwidth}
        \centering
        \includegraphics[width=\linewidth]{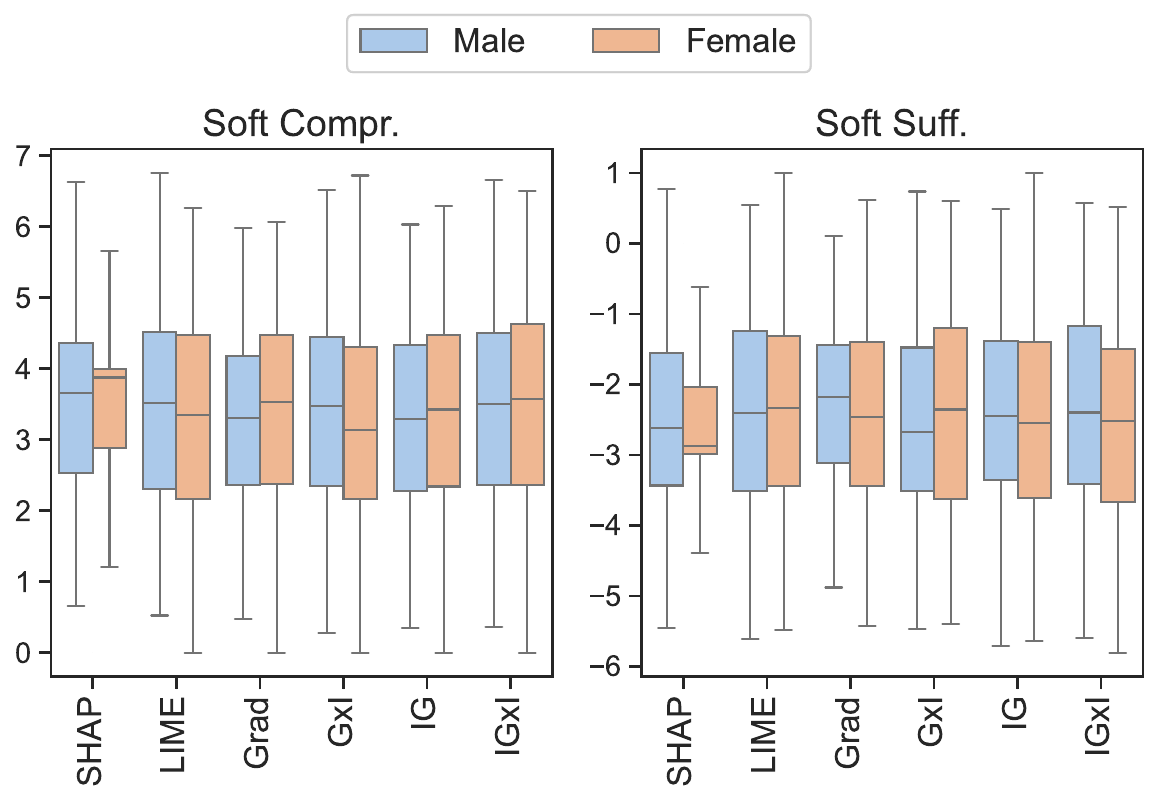}
        \caption{Stereotypes}
        \label{fig:soft_subfig3}
    \end{subfigure}
    \begin{subfigure}{0.45\textwidth}
        \centering
        \includegraphics[width=\linewidth]{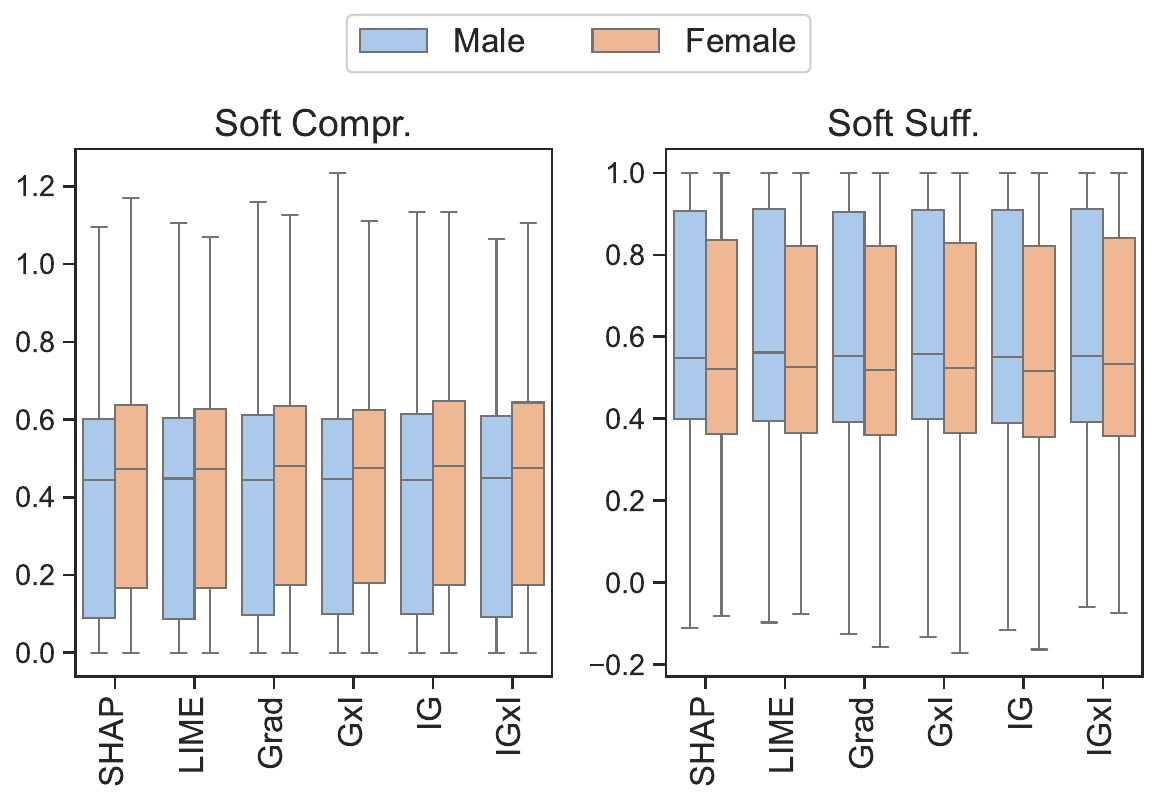}
        \caption{COMPAS}
        \label{fig:soft_subfig4}
    \end{subfigure}
    \caption{\textbf{Box-plots of the soft comprehensiveness and sufficiency metrics} obtained over 5 runs for each using TinyBERT on GECO-ALL, Stereotypes, and COMPAS, including the runs not resulting in statistically significant disparity.}
    \label{fig:soft_boxplots}
\end{figure*}

\begin{figure*}[t]
    \centering
    \begin{subfigure}{\textwidth}
        \centering
        \includegraphics[width=\linewidth]{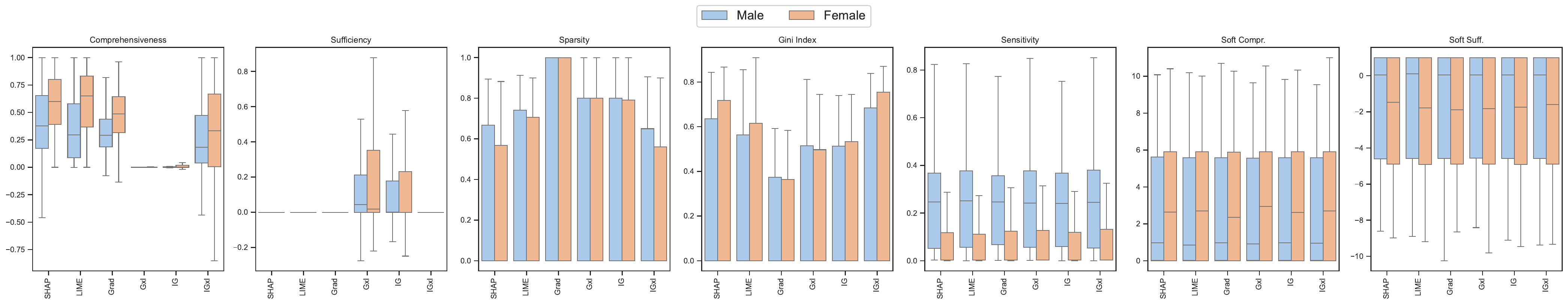}
        \caption{GECO-ALL}
        \label{fig:subfig1}
    \end{subfigure}

    \begin{subfigure}{\textwidth}
        \centering
        \includegraphics[width=\linewidth]{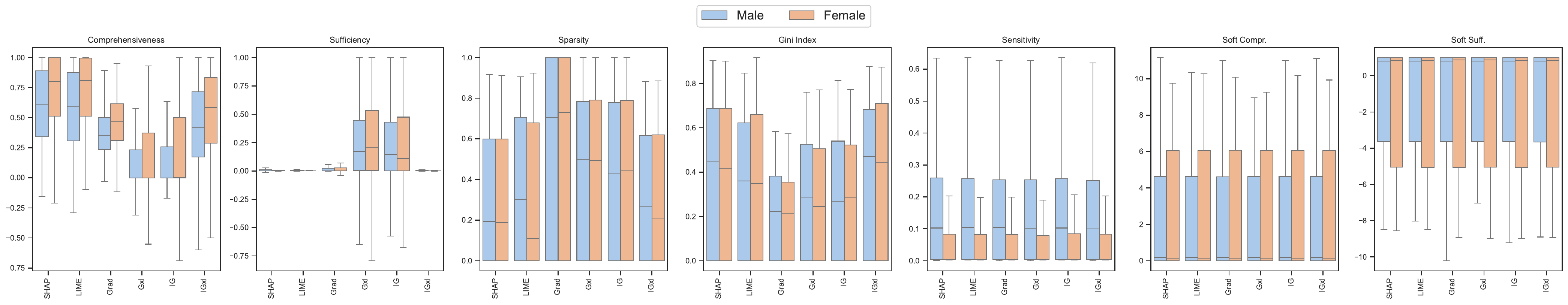}
        \caption{GECO-SUBJ}
        \label{fig:subfig1}
    \end{subfigure}
    
    \begin{subfigure}{\textwidth}
        \centering
        \includegraphics[width=\linewidth]{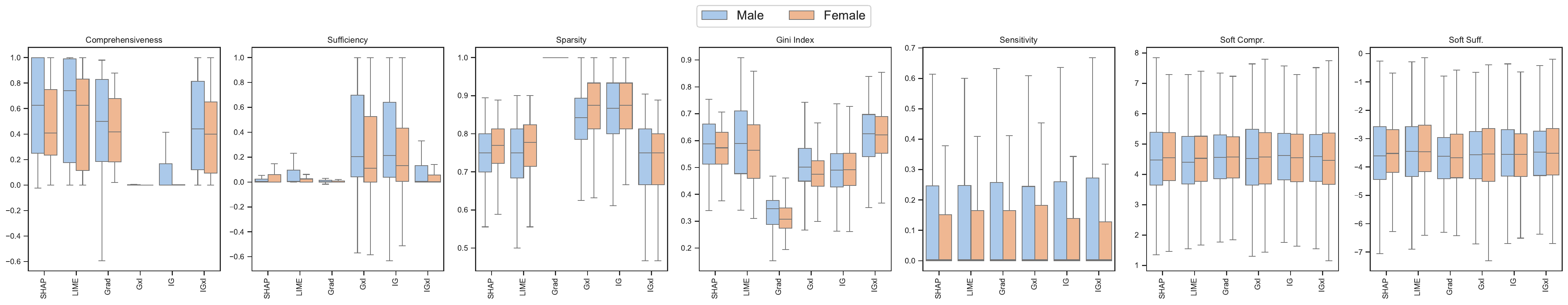}
        \caption{Stereotypes}
        \label{fig:subfig3}
    \end{subfigure}

    \begin{subfigure}{\textwidth}
        \centering
        \includegraphics[width=\linewidth]{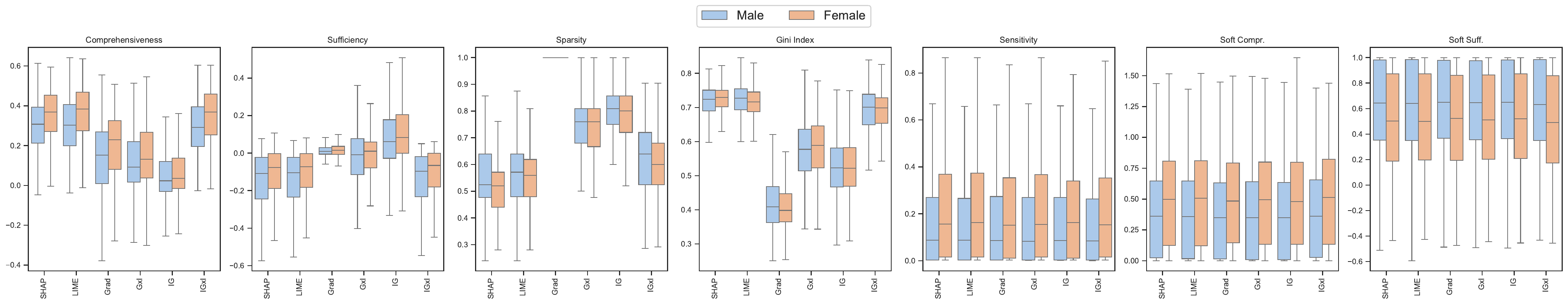}
        \caption{COMPAS}
        \label{fig:subfig4}
    \end{subfigure}
    \caption{\textbf{Box-plots of evaluation scores} obtained over 5 runs for each using BERT on our datasets, including the runs not resulting in statistically significant disparity.}
    \label{fig:boxplots_bert}
\end{figure*}

\begin{figure*}[t]
    \centering
    \begin{subfigure}{\textwidth}
        \centering
        \includegraphics[width=\linewidth]{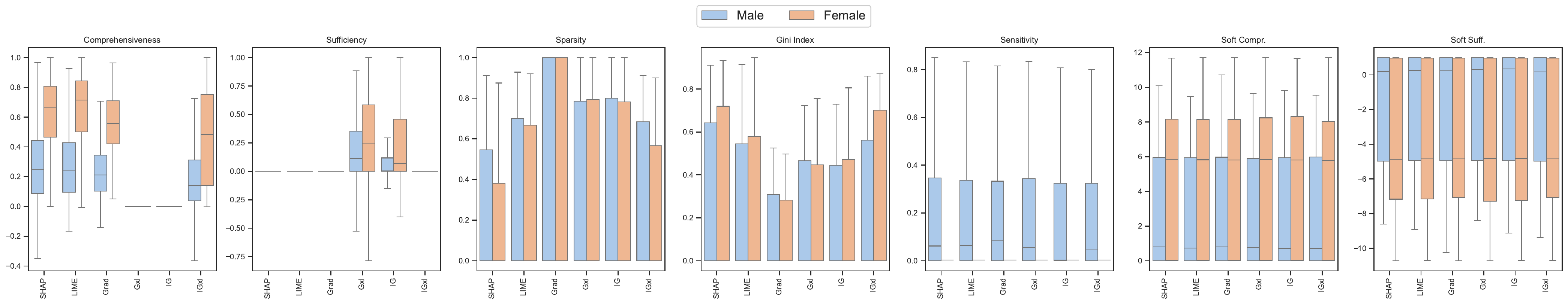}
        \caption{GECO-ALL}
        \label{fig:subfig1}
    \end{subfigure}

    \begin{subfigure}{\textwidth}
        \centering
        \includegraphics[width=\linewidth]{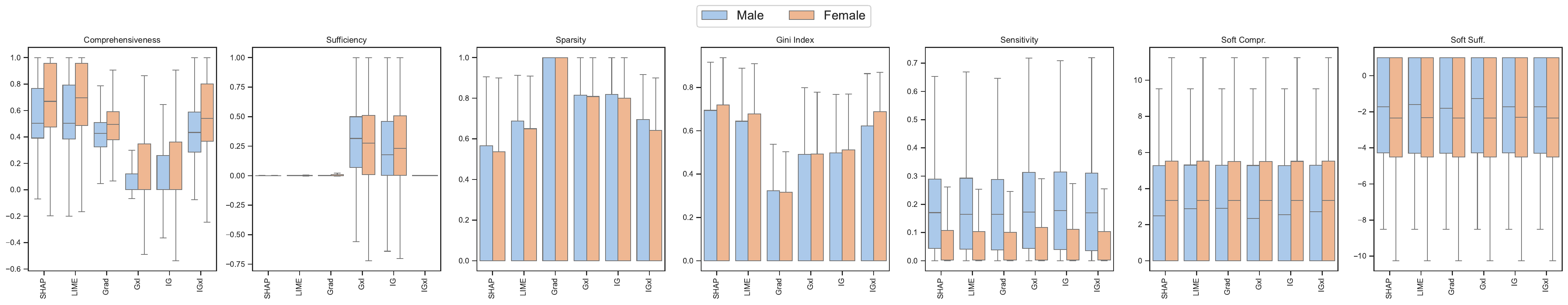}
        \caption{GECO-SUBJ}
        \label{fig:subfig1}
    \end{subfigure}
    
    \begin{subfigure}{\textwidth}
        \centering
        \includegraphics[width=\linewidth]{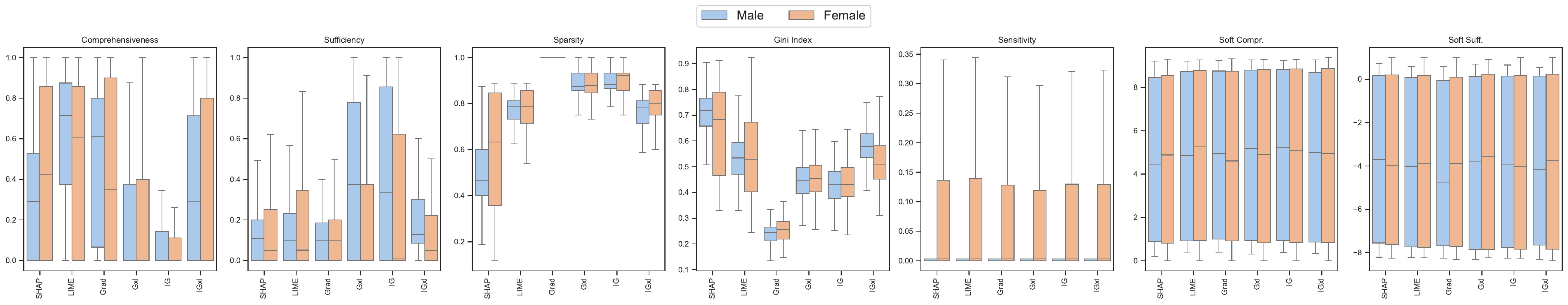}
        \caption{Stereotypes}
        \label{fig:subfig3}
    \end{subfigure}

    \begin{subfigure}{\textwidth}
        \centering
        \includegraphics[width=\linewidth]{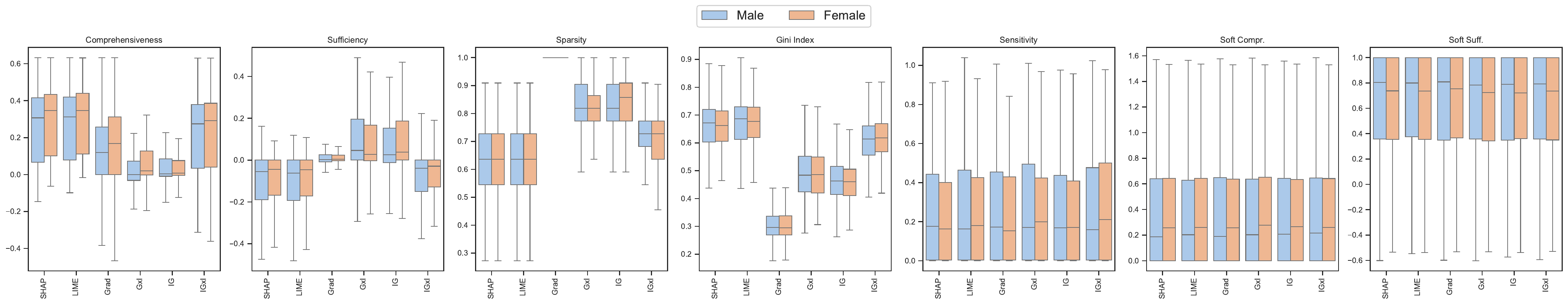}
        \caption{COMPAS}
        \label{fig:subfig4}
    \end{subfigure}
    \caption{\textbf{Box-plots of evaluation scores} obtained over 5 runs for each using FairBERTa on our datasets, including the runs not resulting in statistically significant disparity.}
    \label{fig:boxplots_fairberta}
\end{figure*}

\begin{figure*}[t]
    \centering
    \begin{subfigure}{\textwidth}
        \centering
        \includegraphics[width=\linewidth]{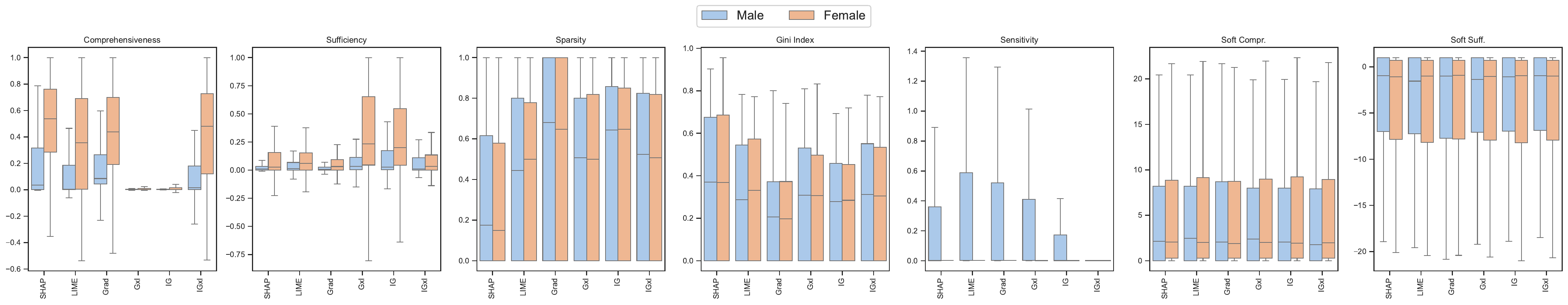}
        \caption{GECO-ALL}
        \label{fig:subfig1}
    \end{subfigure}

    \begin{subfigure}{\textwidth}
        \centering
        \includegraphics[width=\linewidth]{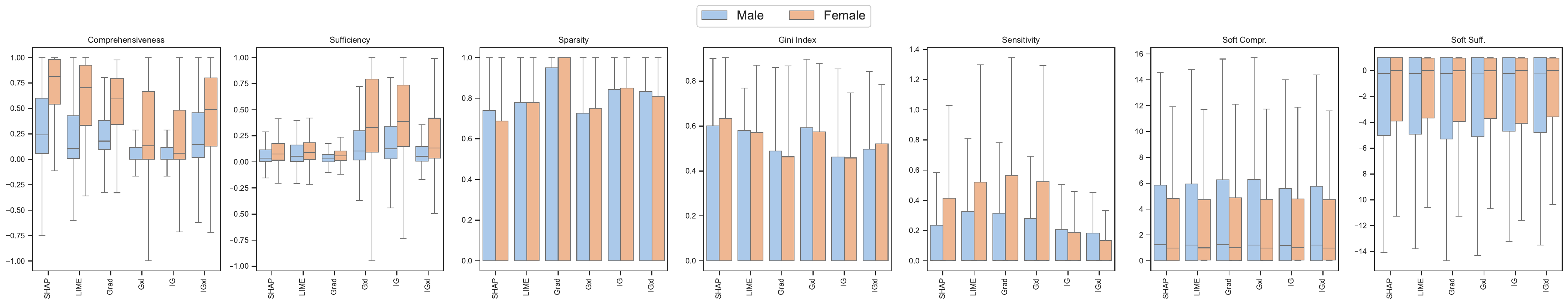}
        \caption{GECO-SUBJ}
        \label{fig:subfig1}
    \end{subfigure}
    
    \begin{subfigure}{\textwidth}
        \centering
        \includegraphics[width=\linewidth]{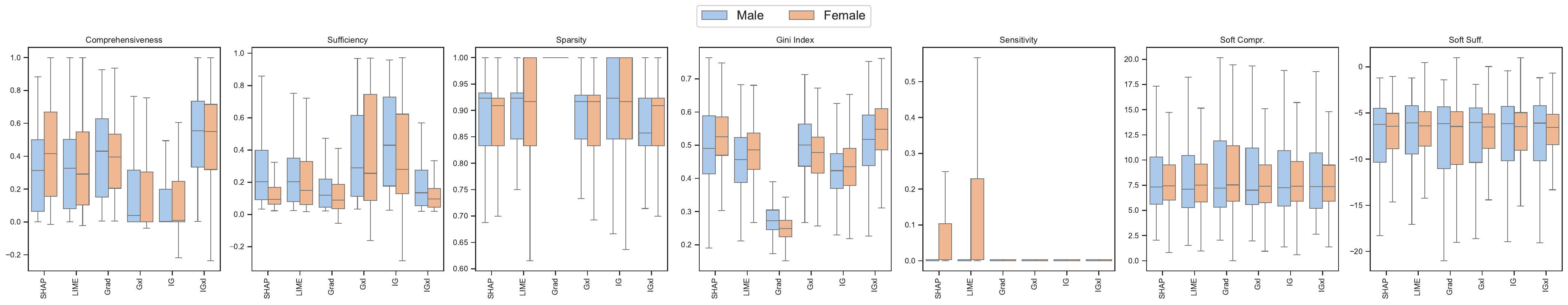}
        \caption{Stereotypes}
        \label{fig:subfig3}
    \end{subfigure}

    \begin{subfigure}{\textwidth}
        \centering
        \includegraphics[width=\linewidth]{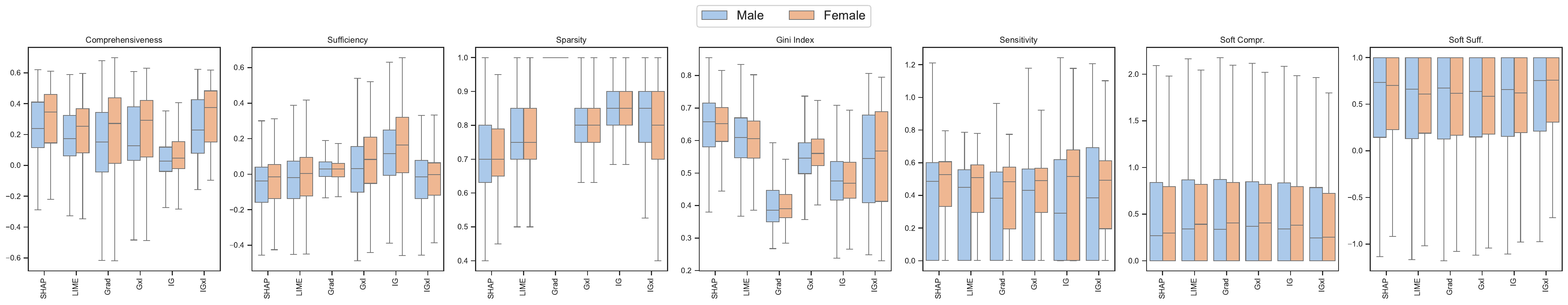}
        \caption{COMPAS}
        \label{fig:subfig4}
    \end{subfigure}
    \caption{\textbf{Box-plots of evaluation scores} obtained over 5 runs for each using GPT-2 on our datasets, including the runs not resulting in statistically significant disparity.}
    \label{fig:boxplots_gpt2}
\end{figure*}

\begin{figure*}[t]
    \centering
    \begin{subfigure}{\textwidth}
        \centering
        \includegraphics[width=\linewidth]{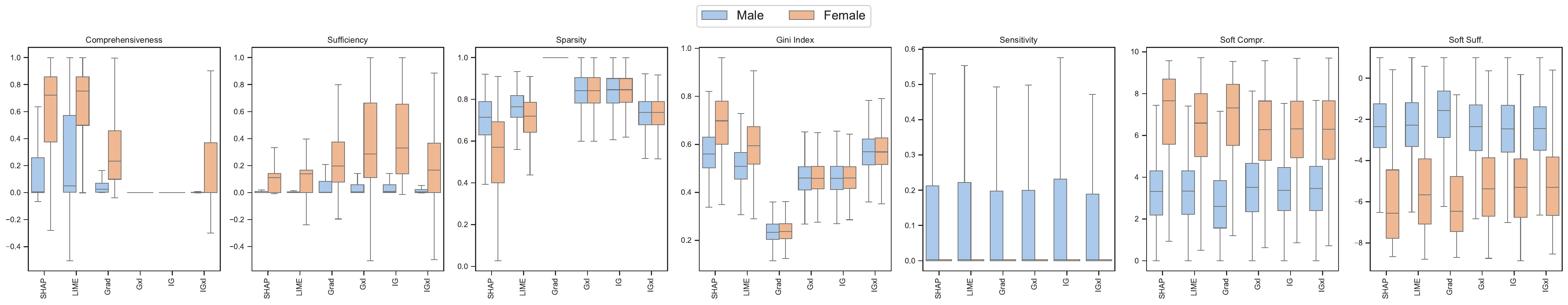}
        \caption{GECO-ALL}
        \label{fig:subfig1}
    \end{subfigure}

    \begin{subfigure}{\textwidth}
        \centering
        \includegraphics[width=\linewidth]{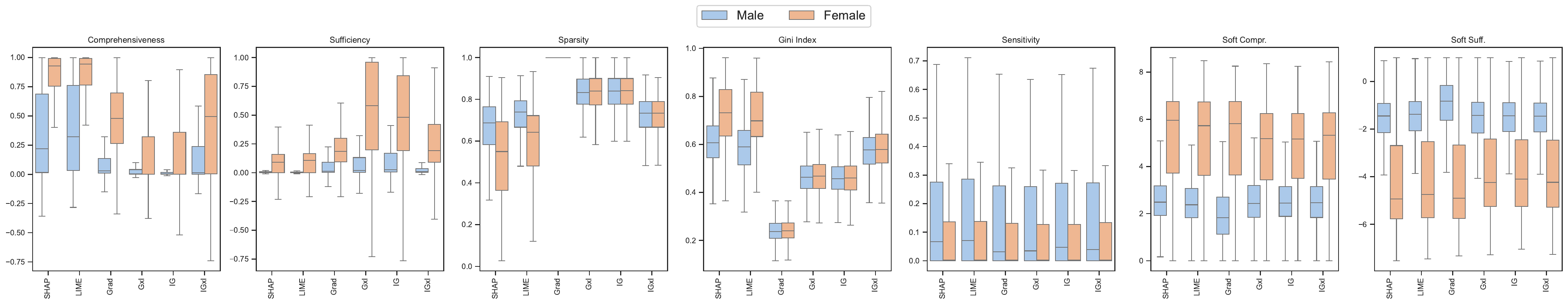}
        \caption{GECO-SUBJ}
        \label{fig:subfig1}
    \end{subfigure}
    
    \begin{subfigure}{\textwidth}
        \centering
        \includegraphics[width=\linewidth]{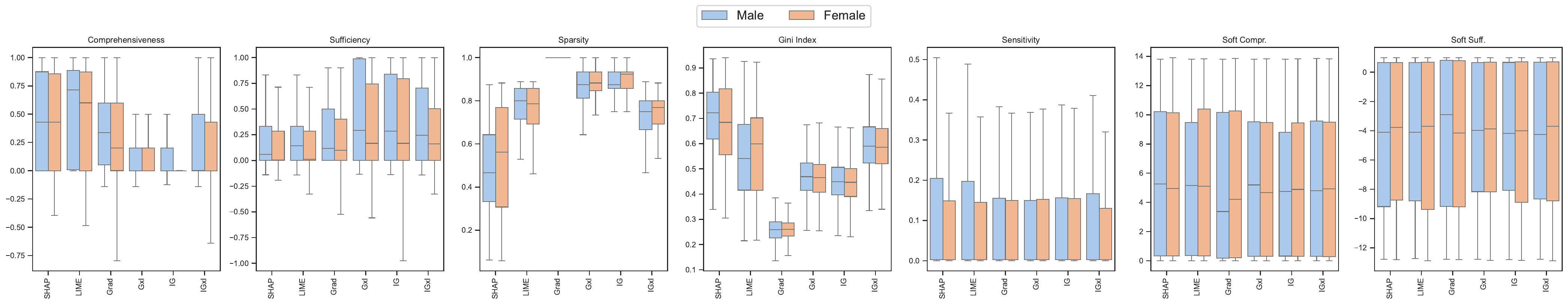}
        \caption{Stereotypes}
        \label{fig:subfig3}
    \end{subfigure}

    \begin{subfigure}{\textwidth}
        \centering
        \includegraphics[width=\linewidth]{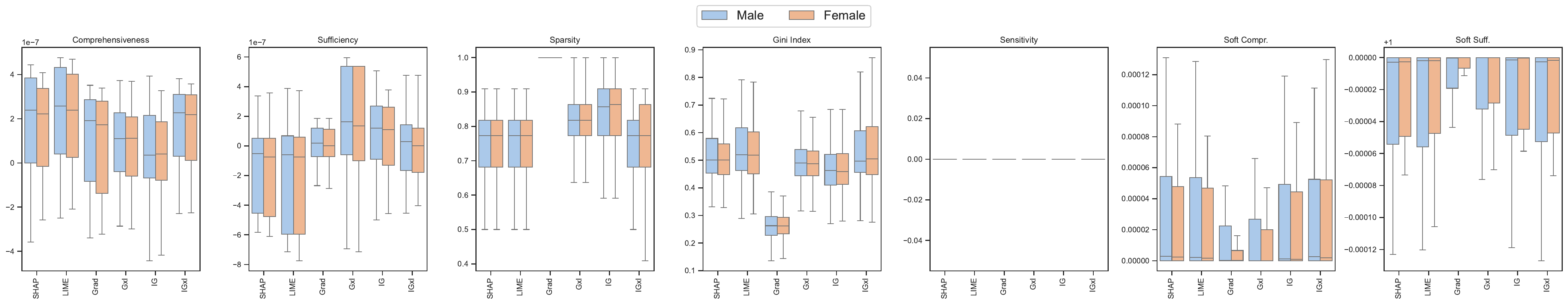}
        \caption{COMPAS}
        \label{fig:subfig4}
    \end{subfigure}
    \caption{\textbf{Box-plots of evaluation scores} obtained over 5 runs for each using RoBERTa on our datasets, including the runs not resulting in statistically significant disparity.}
    \label{fig:boxplots_roberta}
\end{figure*}

We display further results from our experiments in the tables and figures below. Tables \ref{tab:full_geco_all}, \ref{tab:full_geco_subj}, \ref{tab:full_compas}, and \ref{tab:full_stereo} display further results on the number of runs resulting in significant disparity as well as the average effect sizes. Figures \ref{fig:soft_boxplots}, \ref{fig:boxplots_bert}, \ref{fig:boxplots_fairberta}, \ref{fig:boxplots_gpt2}, and \ref{fig:boxplots_roberta} display box plots of the distributions of all evaluation scores obtained over all runs, including the ones not resulting in significant disparity for the remainder of our models. 

\begin{landscape}
\begin{table}[t]
\small
\centering
\caption{\label{tab:full_geco_all}\textbf{Occurence of disparity and effect sizes on GECO-ALL}. The numbers in parentheses display how many of the 5 runs resulted in statistically significant disparity, along with the effect sizes (Cohen's $d$). Bold font indicates considerable effect size ($\vert d \vert \geq 0.2$). }
\begin{tabular}{lllllllll}
    \toprule
    \textbf{Model} & 
    \textbf{Method} & 
    AOPC Compr. $(\uparrow)$ & 
    AOPC Suff. $(\downarrow)$ & 
    Soft Compr. $(\uparrow)$ & 
    Soft Suff.  $(\downarrow)$ & 
    Gini Index $(\uparrow)$ & 
    Sparsity $(\downarrow)$ & 
    Sens. $(\downarrow)$ 
    \\
\midrule
\multirow[t]{6}{*}{BERT} & Grad & \textbf{(5) -1.68±1.28} & (4) .16±.22 & \textbf{(5) .60±1.32} & (0) NA & \textbf{(5) 1.00±1.92} & (5) -.06±.19 & \textbf{(5) -.61±1.41} \\
 & GxI & \textbf{(4) -.22±.29} & \textbf{(4) .27±.05} & \textbf{(5) .74±.95} & (0) NA & \textbf{(5) 1.06±1.89} & \textbf{(4) -.34±.49} & \textbf{(5) -.77±.92} \\
 & IG & \textbf{(4) -.31±.35} & (5) -.11±.19 & \textbf{(4) .88±1.15} & \textbf{(1) .24±.00} & \textbf{(5) 1.09±1.93} & \textbf{(4) -.27±.36} & \textbf{(4) -.87±1.19} \\
 & IGxI & \textbf{(5) -.50±.78} & \textbf{(5) -1.12±.23} & \textbf{(4) 1.18±1.97} & \textbf{(5) .74±.22} & \textbf{(5) 1.02±1.99} & (5) -.11±.06 & \textbf{(4) -1.08±1.97} \\
 & LIME & \textbf{(5) -1.46±1.12} & \textbf{(5) -.61±.54} & \textbf{(5) .73±.95} & \textbf{(4) .59±.13} & \textbf{(5) 1.08±1.93} & (5) .04±.05 & \textbf{(5) -.71±.97} \\
 & SHAP & \textbf{(5) -1.42±1.38} & \textbf{(5) -1.01±1.78} & \textbf{(5) .67±.90} & \textbf{(5) .72±1.15} & \textbf{(5) .98±1.93} & (5) -.04±.02 & \textbf{(4) -.84±.95} \\
\cmidrule{1-9}
\multirow[t]{6}{*}{FairBERTa} & Grad & \textbf{(5) -2.41±.99} & \textbf{(4) .45±.21} & \textbf{(5) 17.86±31.01} & (0) NA & \textbf{(4) 1.00±.58} & \textbf{(4) -.30±.25} & \textbf{(5) -15.63±26.83} \\
 & GxI & \textbf{(3) -.47±.09} & \textbf{(2) .28±.12} & \textbf{(5) 19.77±35.04} & (0) NA & \textbf{(4) .97±.56} & \textbf{(4) -.61±.56} & \textbf{(5) -11.54±18.47} \\
 & IG & \textbf{(3) -.32±.19} & \textbf{(4) -.42±.13} & \textbf{(5) 10.82±16.97} & \textbf{(3) .22±.07} & \textbf{(4) .91±.59} & \textbf{(4) -.69±.27} & \textbf{(5) -9.99±15.34} \\
 & IGxI & \textbf{(5) -.97±.22} & \textbf{(4) -1.90±.94} & \textbf{(5) 13.73±22.83} & \textbf{(4) 1.05±.51} & \textbf{(4) .94±.56} & \textbf{(4) -.31±.28} & \textbf{(5) -23.68±42.54} \\
 & LIME & \textbf{(5) -1.65±.45} & \textbf{(3) -.60±.25} & \textbf{(5) 20.88±37.26} & \textbf{(3) .46±.14} & \textbf{(4) .99±.58} & \textbf{(4) -.28±.22} & \textbf{(5) -19.44±34.23} \\
 & SHAP & \textbf{(5) -1.52±.38} & \textbf{(3) -1.37±.91} & \textbf{(5) 15.23±25.81} & \textbf{(3) 1.46±.94} & \textbf{(4) 1.01±.62} & \textbf{(4) -.29±.25} & \textbf{(5) -10.53±16.58} \\
\cmidrule{1-9}
\multirow[t]{6}{*}{GPT2} & Grad & \textbf{(5) -.52±3.28} & (3) -.06±.66 & (5) .06±1.32 & \textbf{(3) .24±.48} & \textbf{(5) .58±1.90} & \textbf{(5) -.30±1.08} & (5) -.04±1.28 \\
 & GxI & (5) -.05±.84 & \textbf{(5) .24±.20} & (4) .19±1.24 & \textbf{(1) -.39±.00} & \textbf{(5) .51±1.26} & \textbf{(5) -.37±1.43} & \textbf{(4) -.24±1.26} \\
 & IG & \textbf{(3) .33±.71} & (1) .15±.00 & \textbf{(5) .28±1.11} & (0) NA & \textbf{(5) .44±.81} & \textbf{(5) -.32±1.48} & \textbf{(4) -.32±1.26} \\
 & IGxI & \textbf{(5) -1.03±.90} & (5) .07±.67 & \textbf{(5) .33±1.19} & (4) .14±.47 & \textbf{(5) .26±.51} & \textbf{(5) .24±1.40} & \textbf{(5) -.32±1.23} \\
 & LIME & \textbf{(5) -.60±.88} & \textbf{(5) -.27±.61} & (5) .13±1.26 & \textbf{(2) .55±.18} & \textbf{(5) 1.06±1.92} & (5) -.19±1.15 & \textbf{(5) -.20±1.34} \\
 & SHAP & \textbf{(5) -.64±1.00} & (5) -.05±.18 & \textbf{(5) .22±1.30} & (5) .15±.37 & \textbf{(5) .42±1.25} & \textbf{(5) -.32±1.07} & (5) -.19±1.39 \\
\cmidrule{1-9}
\multirow[t]{6}{*}{TinyBERT} & Grad & \textbf{(5) -.84±1.02} & \textbf{(4) -.55±.21} & \textbf{(5) 3.47±3.67} & \textbf{(3) .35±.05} & \textbf{(5) -1.23±1.16} & (5) .04±.05 & \textbf{(5) -3.37±3.70} \\
 & GxI & \textbf{(4) -.54±.02} & \textbf{(5) -.77±.13} & \textbf{(5) 3.40±3.14} & \textbf{(5) .61±.14} & \textbf{(5) -1.21±1.17} & (5) .09±.32 & \textbf{(4) -4.02±3.01} \\
 & IG & \textbf{(4) -.45±.51} & \textbf{(2) -.91±.53} & \textbf{(4) 4.51±3.22} & \textbf{(2) .64±.34} & \textbf{(5) -1.22±1.16} & \textbf{(5) .27±.41} & \textbf{(4) -4.25±3.11} \\
 & IGxI & \textbf{(5) -.29±.44} & \textbf{(5) -1.17±.53} & \textbf{(5) 3.25±3.95} & \textbf{(5) 1.23±.46} & \textbf{(5) -1.25±1.18} & (4) -.03±.02 & \textbf{(5) -3.66±4.37} \\
 & LIME & \textbf{(5) -.61±.44} & \textbf{(5) -1.24±.28} & \textbf{(5) 2.98±3.05} & \textbf{(5) .75±.15} & \textbf{(5) -1.22±1.17} & (4) .00±.05 & \textbf{(5) -2.93±3.01} \\
 & SHAP & \textbf{(4) -1.11±.14} & \textbf{(5) -2.01±.29} & \textbf{(5) 3.36±3.55} & \textbf{(5) 2.00±.26} & \textbf{(5) -1.18±1.16} & (3) .07±.04 & \textbf{(5) -3.46±3.77} \\
\cmidrule{1-9}
\multirow[t]{6}{*}{RoBERTa} & Grad & \textbf{(5) -1.25±1.67} & (0) NA & \textbf{(5) 3.66±2.87} & (0) NA & \textbf{(5) .48±.85} & \textbf{(5) -1.12±1.64} & \textbf{(5) -3.36±2.57} \\
 & GxI & \textbf{(2) -.50±.12} & (2) -.01±.18 & \textbf{(5) 2.27±2.08} & (1) .18±.00 & \textbf{(5) .44±.90} & \textbf{(5) -1.12±1.56} & \textbf{(5) -2.00±2.00} \\
 & IG & (3) -.18±.49 & (0) NA & \textbf{(5) 2.17±2.03} & (0) NA & (5) -.20±1.93 & \textbf{(5) -1.18±1.69} & \textbf{(5) -2.22±1.96} \\
 & IGxI & \textbf{(4) -.53±.83} & \textbf{(1) -.28±.00} & \textbf{(5) 2.12±2.03} & (0) NA & \textbf{(5) .40±.97} & \textbf{(5) -.95±1.37} & \textbf{(5) -2.09±2.02} \\
 & LIME & \textbf{(5) -2.68±3.36} & \textbf{(5) -.88±1.16} & \textbf{(5) 2.56±2.25} & \textbf{(5) .49±.63} & \textbf{(5) .49±.91} & \textbf{(5) -.90±.99} & \textbf{(5) -2.52±2.14} \\
 & SHAP & \textbf{(5) -2.23±2.18} & \textbf{(5) -1.42±1.29} & \textbf{(4) 3.76±2.15} & \textbf{(5) 1.23±1.23} & \textbf{(5) .54±.89} & \textbf{(5) -.72±.94} & \textbf{(4) -4.05±2.31} \\
\bottomrule
\end{tabular}
\end{table}
\end{landscape}

\begin{landscape}
\begin{table}[t]
\small
\centering
\caption{\label{tab:full_geco_subj}\textbf{Occurence of disparity and effect sizes on GECO-SUBJ}. The numbers in parentheses display how many of the 5 runs resulted in statistically significant disparity, along with the effect sizes (Cohen's $d$). Bold font indicates considerable effect size ($\vert d \vert \geq 0.2$). }
\begin{tabular}{lllllllll}
    \toprule
    \textbf{Model} & 
    \textbf{Method} & 
    AOPC Compr. $(\uparrow)$ & 
    AOPC Suff. $(\downarrow)$ & 
    Soft Compr. $(\uparrow)$ & 
    Soft Suff.  $(\downarrow)$ & 
    Gini Index $(\uparrow)$ & 
    Sparsity $(\downarrow)$ & 
    Sens. $(\downarrow)$ 
    \\
\midrule
\multirow[t]{6}{*}{BERT} & Grad & (5) -.18±.43 & \textbf{(5) .37±.17} & \textbf{(5) 1.54±1.22} & (0) NA & \textbf{(4) .74±.39} & (5) -.06±.11 & \textbf{(5) -1.60±1.29} \\
 & GxI & \textbf{(4) -.23±.25} & \textbf{(4) .30±.06} & \textbf{(5) 1.69±1.43} & (0) NA & \textbf{(3) .96±.19} & \textbf{(2) -.40±.43} & \textbf{(5) -1.71±1.50} \\
 & IG & (5) -.17±.35 & (3) .18±.05 & \textbf{(5) 1.62±1.32} & (0) NA & \textbf{(4) .77±.41} & \textbf{(1) -.30±.00} & \textbf{(5) -1.62±1.37} \\
 & IGxI & \textbf{(5) -.25±.41} & \textbf{(5) -.36±.10} & \textbf{(5) 1.47±1.30} & (0) NA & \textbf{(3) .94±.24} & (4) .06±.08 & \textbf{(5) -1.58±1.33} \\
 & LIME & \textbf{(5) -.32±.54} & \textbf{(3) -.63±.16} & \textbf{(5) 1.61±1.30} & \textbf{(3) .48±.04} & \textbf{(4) .76±.40} & (3) .02±.10 & \textbf{(5) -1.53±1.25} \\
 & SHAP & \textbf{(5) -.30±.58} & (4) -.11±.58 & \textbf{(5) 1.61±1.34} & (4) .10±.39 & \textbf{(4) .75±.40} & (2) .05±.09 & \textbf{(5) -1.56±1.32} \\
\cmidrule{1-9}
\multirow[t]{6}{*}{FairBERTa} & Grad & \textbf{(3) -.59±.23} & (4) .08±.62 & \textbf{(5) -.21±1.07} & (0) NA & \textbf{(4) 1.02±.30} & (5) .04±.10 & \textbf{(5) .20±1.08} \\
 & GxI & \textbf{(3) -.52±.18} & (3) -.10±.27 & (5) -.15±1.10 & (1) .16±.00 & \textbf{(4) 1.02±.29} & (3) .11±.45 & (5) .18±1.08 \\
 & IG & (4) -.06±.24 & \textbf{(2) -.25±.03} & (5) -.17±1.09 & (2) .19±.04 & \textbf{(4) 1.02±.26} & \textbf{(3) -.31±.04} & (5) .17±1.09 \\
 & IGxI & \textbf{(4) -.55±.17} & \textbf{(5) -.72±.37} & (5) -.20±1.08 & \textbf{(5) .50±.24} & \textbf{(4) 1.03±.23} & (3) .13±.07 & (5) .20±1.08 \\
 & LIME & \textbf{(5) -.64±.29} & \textbf{(4) -.54±.20} & (5) -.20±1.08 & \textbf{(4) .47±.15} & \textbf{(4) 1.03±.25} & (3) .05±.09 & (5) .20±1.08 \\
 & SHAP & \textbf{(5) -.55±.28} & \textbf{(3) -.32±.07} & (5) -.20±1.07 & \textbf{(3) .28±.09} & \textbf{(4) 1.01±.27} & (4) .01±.09 & (5) .17±1.08 \\
\cmidrule{1-9}
\multirow[t]{6}{*}{GPT2} & Grad & \textbf{(5) -1.16±1.87} & \textbf{(2) .73±.30} & \textbf{(5) .29±3.90} & \textbf{(3) -.37±.42} & \textbf{(5) -.46±2.33} & \textbf{(4) -.24±1.01} & \textbf{(5) -.34±3.87} \\
 & GxI & \textbf{(4) -.43±.52} & \textbf{(2) .32±.11} & \textbf{(5) .42±3.78} & \textbf{(3) -.26±.12} & (5) -.15±1.79 & \textbf{(5) -.89±1.51} & \textbf{(5) -.47±3.81} \\
 & IG & \textbf{(4) -.62±.77} & (1) .18±.00 & \textbf{(5) .62±3.54} & (0) NA & (5) .01±1.08 & \textbf{(5) -.53±1.24} & \textbf{(5) -.64±3.62} \\
 & IGxI & \textbf{(5) -.47±.93} & \textbf{(5) -.23±.53} & \textbf{(5) .61±3.68} & \textbf{(4) .34±.11} & (5) .05±.93 & \textbf{(4) -.77±1.67} & \textbf{(5) -.57±3.66} \\
 & LIME & \textbf{(4) -1.30±.66} & (4) .08±.47 & \textbf{(5) .52±3.80} & (3) .08±.30 & \textbf{(5) -.40±2.38} & (5) .06±1.00 & \textbf{(5) -.57±3.82} \\
 & SHAP & \textbf{(5) -1.02±1.15} & \textbf{(4) -.55±.59} & \textbf{(5) .49±3.53} & \textbf{(4) .47±.24} & \textbf{(5) -.23±1.90} & (4) -.12±.88 & \textbf{(5) -.51±3.52} \\
\cmidrule{1-9}
\multirow[t]{6}{*}{TinyBERT} & Grad & \textbf{(3) -1.17±.11} & (3) -.11±.19 & \textbf{(5) .82±.72} & (0) NA & \textbf{(4) -.30±.24} & (3) -.01±.02 & \textbf{(5) -.77±.72} \\
 & GxI & \textbf{(5) -.47±.07} & (3) -.16±.28 & \textbf{(5) .75±.69} & (0) NA & \textbf{(4) -.26±.23} & \textbf{(3) .31±.19} & \textbf{(5) -.74±.66} \\
 & IG & \textbf{(5) -.24±.34} & \textbf{(4) -.59±.31} & \textbf{(5) .78±.68} & \textbf{(4) .45±.31} & \textbf{(4) -.26±.26} & \textbf{(4) -.21±.38} & \textbf{(5) -.74±.66} \\
 & IGxI & \textbf{(3) -.71±.07} & \textbf{(5) -1.03±.76} & \textbf{(5) .84±.72} & \textbf{(5) 1.02±.84} & \textbf{(4) -.27±.23} & (5) -.01±.05 & \textbf{(5) -.86±.74} \\
 & LIME & \textbf{(5) -.64±.20} & \textbf{(5) -1.19±.58} & \textbf{(5) .79±.71} & \textbf{(5) .80±.35} & \textbf{(4) -.30±.25} & (5) .05±.05 & \textbf{(5) -.81±.73} \\
 & SHAP & \textbf{(5) -.70±.30} & \textbf{(5) -1.18±.55} & \textbf{(5) .77±.69} & \textbf{(5) .96±.53} & \textbf{(4) -.29±.26} & (5) .04±.04 & \textbf{(5) -.80±.73} \\
\cmidrule{1-9}
\multirow[t]{6}{*}{RoBERTa} & Grad & \textbf{(5) -1.63±.56} & (0) NA & \textbf{(5) 3.10±2.50} & (0) NA & \textbf{(5) .33±.25} & \textbf{(5) -1.01±.67} & \textbf{(5) -3.04±2.57} \\
 & GxI & \textbf{(3) -.28±.22} & (0) NA & \textbf{(5) 2.29±1.90} & (0) NA & \textbf{(5) .35±.24} & \textbf{(5) -1.47±.58} & \textbf{(5) -2.25±1.83} \\
 & IG & \textbf{(1) -.49±.00} & (0) NA & \textbf{(5) 2.31±1.77} & (0) NA & \textbf{(5) .38±.24} & \textbf{(5) -1.28±.33} & \textbf{(5) -2.33±1.81} \\
 & IGxI & \textbf{(4) -1.18±.70} & (2) -.18±.35 & \textbf{(5) 2.33±1.86} & (2) .06±.25 & \textbf{(5) .32±.18} & \textbf{(5) -.95±.12} & \textbf{(5) -2.34±1.83} \\
 & LIME & \textbf{(5) -1.44±.76} & \textbf{(5) -1.25±.60} & \textbf{(5) 2.60±2.02} & \textbf{(5) .90±.34} & \textbf{(5) .38±.19} & \textbf{(4) -.70±.31} & \textbf{(5) -2.65±1.97} \\
 & SHAP & \textbf{(5) -1.62±.75} & \textbf{(5) -1.16±.49} & \textbf{(5) 2.67±2.00} & \textbf{(4) 1.05±.36} & \textbf{(5) .33±.24} & \textbf{(5) -.56±.30} & \textbf{(5) -2.69±1.96} \\
\bottomrule
\end{tabular}
\end{table}
\end{landscape}

\begin{landscape}
\begin{table}[t]
\small
\centering
\caption{\label{tab:full_compas}\textbf{Occurence of disparity and effect sizes on COMPAS}. The numbers in parentheses display how many of the 5 runs resulted in statistically significant disparity, along with the effect sizes (Cohen's $d$). Bold font indicates considerable effect size ($\vert d \vert \geq 0.2$).}
\begin{tabular}{lllllllll}
    \toprule
    \textbf{Model} & 
    \textbf{Method} & 
    AOPC Compr. $(\uparrow)$ & 
    AOPC Suff. $(\downarrow)$ & 
    Soft Compr. $(\uparrow)$ & 
    Soft Suff.  $(\downarrow)$ & 
    Gini Index $(\uparrow)$ & 
    Sparsity $(\downarrow)$ & 
    Sens. $(\downarrow)$ 
    \\
    \midrule
\multirow[t]{6}{*}{BERT} & Grad & \textbf{(5) -.34±.11} & (2) -.17±.01 & \textbf{(3) -.41±.02} & \textbf{(4) .33±.09} & (3) .15±.47 & \textbf{(5) .45±.31} & \textbf{(4) -.37±.18} \\
 & GxI & \textbf{(5) -.35±.51} & (3) .05±.38 & \textbf{(3) -.32±.08} & \textbf{(3) .33±.08} & \textbf{(3) -.87±.47} & \textbf{(3) .80±.54} & \textbf{(2) -.56±.13} \\
 & IG & \textbf{(2) -.31±.17} & \textbf{(1) -.21±.00} & \textbf{(3) -.37±.06} & \textbf{(3) .34±.09} & (4) -.15±.19 & \textbf{(3) .32±.08} & \textbf{(3) -.40±.09} \\
 & IGxI & \textbf{(5) -.45±.12} & \textbf{(5) -.24±.09} & \textbf{(3) -.35±.07} & \textbf{(4) .31±.11} & (2) .11±.43 & (4) .19±.30 & \textbf{(3) -.46±.16} \\
 & LIME & \textbf{(5) -.43±.07} & \textbf{(4) -.27±.09} & \textbf{(3) -.37±.08} & \textbf{(3) .34±.10} & \textbf{(3) .26±.08} & (2) .00±.30 & \textbf{(3) -.47±.16} \\
 & SHAP & \textbf{(5) -.42±.13} & \textbf{(4) -.28±.08} & \textbf{(2) -.39±.05} & \textbf{(3) .33±.10} & (4) -.06±.24 & \textbf{(5) .27±.49} & \textbf{(3) -.45±.16} \\
\cmidrule{1-9}
\multirow[t]{6}{*}{FairBERTa} & Grad & \textbf{(2) -.35±.03} & \textbf{(1) .20±.00} & (2) -.05±.33 & (2) .05±.33 & (4) .10±.24 & (0) NA & (4) -.00±.32 \\
 & GxI & \textbf{(2) -.33±.01} & (3) .12±.25 & (2) -.09±.29 & \textbf{(1) .38±.00} & \textbf{(4) .30±.07} & (3) -.19±.01 & (4) .06±.38 \\
 & IG & (5) .12±.23 & \textbf{(4) -.29±.13} & (2) -.08±.30 & (2) .06±.32 & (4) .06±.27 & (2) .08±.25 & (2) -.08±.28 \\
 & IGxI & (3) .11±.34 & \textbf{(2) -.37±.04} & \textbf{(1) -.38±.00} & (2) .08±.30 & (2) .15±.54 & (4) .15±.30 & \textbf{(3) -.21±.32} \\
 & LIME & \textbf{(1) -.37±.00} & \textbf{(1) -.40±.00} & \textbf{(1) -.38±.00} & \textbf{(1) .38±.00} & (5) .11±.22 & (2) -.02±.30 & (3) .04±.35 \\
 & SHAP & \textbf{(1) -.37±.00} & \textbf{(1) -.36±.00} & (2) -.08±.30 & (2) .08±.30 & (4) .18±.29 & (2) -.10±.41 & (3) .05±.39 \\
\cmidrule{1-9}
\multirow[t]{6}{*}{GPT2} & Grad & \textbf{(5) -.43±.17} & \textbf{(4) .33±.06} & \textbf{(1) -.26±.00} & (4) -.02±.13 & (3) .00±.18 & \textbf{(1) .21±.00} & \textbf{(4) -.39±.03} \\
 & GxI & \textbf{(5) -.34±.10} & \textbf{(4) -.41±.52} & (3) .10±.28 & \textbf{(2) -.22±.15} & \textbf{(5) -.22±.10} & (4) -.02±.39 & \textbf{(4) -.37±.04} \\
 & IG & (4) -.14±.06 & \textbf{(4) -.30±.39} & (5) .08±.18 & (5) -.09±.17 & \textbf{(1) .49±.00} & (4) -.03±.23 & \textbf{(3) -.46±.12} \\
 & IGxI & \textbf{(4) -.54±.10} & (3) .10±.21 & (5) .07±.18 & (5) -.08±.19 & \textbf{(5) -.23±.54} & \textbf{(4) .35±.49} & \textbf{(1) -.31±.00} \\
 & LIME & \textbf{(5) -.48±.05} & \textbf{(2) -.26±.00} & \textbf{(1) .35±.00} & (4) -.15±.11 & \textbf{(4) .24±.28} & \textbf{(3) -.27±.05} & \textbf{(4) -.41±.03} \\
 & SHAP & \textbf{(4) -.44±.03} & (3) -.17±.29 & (0) NA & (0) NA & \textbf{(5) .21±.21} & \textbf{(4) -.55±.06} & \textbf{(4) -.40±.06} \\
\cmidrule{1-9}
\multirow[t]{6}{*}{TinyBERT} & Grad & \textbf{(4) .40±.39} & (4) -.06±.31 & \textbf{(1) -.51±.00} & \textbf{(1) .49±.00} & \textbf{(2) .40±.20} & (5) .02±.26 & \textbf{(4) -.50±.12} \\
 & GxI & \textbf{(5) .53±.05} & \textbf{(2) -.51±.20} & \textbf{(1) -.48±.00} & \textbf{(1) .50±.00} & \textbf{(5) .59±.19} & \textbf{(5) -.40±.08} & \textbf{(4) -.53±.11} \\
 & IG & \textbf{(1) .40±.00} & (5) .18±.30 & \textbf{(1) -.46±.00} & \textbf{(1) .49±.00} & (2) .19±.39 & (2) .03±.23 & \textbf{(5) -.44±.07} \\
 & IGxI & \textbf{(4) .32±.26} & (4) .05±.22 & \textbf{(1) -.52±.00} & \textbf{(1) .51±.00} & \textbf{(5) .59±.18} & \textbf{(4) -.66±.10} & \textbf{(5) -.48±.15} \\
 & LIME & (5) .14±.44 & (4) .03±.22 & \textbf{(1) -.51±.00} & \textbf{(1) .53±.00} & \textbf{(4) .29±.10} & \textbf{(1) -.38±.00} & \textbf{(3) -.51±.05} \\
 & SHAP & \textbf{(3) .69±.00} & (4) .01±.22 & \textbf{(1) -.49±.00} & \textbf{(1) .50±.00} & \textbf{(4) .28±.07} & (5) -.09±.20 & \textbf{(5) -.51±.10} \\
\cmidrule{1-9}
\multirow[t]{6}{*}{RoBERTa} & Grad & (3) .11±.32 & (2) -.04±.18 & \textbf{(1) -.33±.00} & \textbf{(2) .27±.06} & (3) -.08±.26 & (0) NA & (0) NA \\
 & GxI & (1) .07±.00 & (2) -.07±.28 & (2) -.13±.19 & \textbf{(1) .33±.00} & \textbf{(1) .41±.00} & \textbf{(1) -.32±.00} & (0) NA \\
 & IG & \textbf{(1) -.48±.00} & (3) .08±.24 & \textbf{(1) -.33±.00} & \textbf{(1) .33±.00} & \textbf{(1) -.34±.00} & \textbf{(1) -.21±.00} & (0) NA \\
 & IGxI & (3) .00±.22 & \textbf{(3) .23±.27} & \textbf{(1) -.33±.00} & (2) .14±.18 & \textbf{(3) -.21±.41} & (3) -.00±.41 & (0) NA \\
 & LIME & \textbf{(1) -.32±.00} & (1) .10±.00 & \textbf{(1) -.33±.00} & \textbf{(1) .33±.00} & (1) -.11±.00 & (0) NA & (0) NA \\
 & SHAP & (2) .03±.35 & (3) .17±.36 & \textbf{(1) -.33±.00} & \textbf{(1) .33±.00} & \textbf{(2) .27±.02} & (4) -.14±.27 & (0) NA \\
\bottomrule
\end{tabular}
\end{table}
\end{landscape}

\begin{landscape}
\begin{table}[t]
\small
\centering
\caption{\label{tab:full_stereo}\textbf{Occurence of disparity and effect sizes on Stereotypes}. The numbers in parentheses display how many of the 5 runs resulted in statistically significant disparity, along with the effect sizes (Cohen's $d$). Bold font indicates considerable effect size ($\vert d \vert \geq 0.2$). }
\begin{tabular}{lllllllll}
    \toprule
    \textbf{Model} & 
    \textbf{Method} & 
    AOPC Compr. $(\uparrow)$ & 
    AOPC Suff. $(\downarrow)$ & 
    Soft Compr. $(\uparrow)$ & 
    Soft Suff.  $(\downarrow)$ & 
    Gini Index $(\uparrow)$ & 
    Sparsity $(\downarrow)$ & 
    Sens. $(\downarrow)$ 
    \\
\midrule
\multirow[t]{6}{*}{BERT} & Grad & \textbf{(5) .25±.18} & (5) -.13±.23 & (0) NA & (0) NA & \textbf{(5) .32±.04} & (0) NA & \textbf{(3) .24±.29} \\
 & GxI & \textbf{(2) .51±.00} & \textbf{(2) .54±.00} & (0) NA & (0) NA & \textbf{(3) .54±.00} & \textbf{(5) -.34±.11} & \textbf{(2) .24±.34} \\
 & IG & \textbf{(5) .35±.06} & \textbf{(5) .22±.11} & (0) NA & (0) NA & (0) NA & (0) NA & \textbf{(4) .29±.16} \\
 & IGxI & \textbf{(2) .43±.00} & \textbf{(5) .27±.48} & (0) NA & (0) NA & (0) NA & (0) NA & \textbf{(4) .33±.16} \\
 & LIME & \textbf{(2) .53±.00} & \textbf{(5) .29±.30} & (0) NA & (0) NA & \textbf{(5) .33±.15} & \textbf{(2) -.55±.00} & \textbf{(2) .33±.21} \\
 & SHAP & \textbf{(5) .36±.17} & (5) -.11±.55 & (0) NA & (0) NA & \textbf{(2) .38±.00} & \textbf{(5) -.36±.08} & (5) .19±.20 \\
\cmidrule{1-9}
\multirow[t]{6}{*}{FairBERTa} & Grad & (0) NA & (0) NA & (0) NA & (0) NA & \textbf{(5) -.38±.00} & (0) NA & \textbf{(1) -.38±.00} \\
 & GxI & (0) NA & \textbf{(5) .58±.00} & (0) NA & (0) NA & (0) NA & (0) NA & \textbf{(1) -.48±.00} \\
 & IG & (0) NA & \textbf{(5) .30±.00} & (0) NA & (0) NA & (0) NA & (0) NA & \textbf{(3) -.37±.03} \\
 & IGxI & (0) NA & \textbf{(5) .41±.00} & (0) NA & (0) NA & \textbf{(5) .78±.00} & \textbf{(5) -.37±.00} & (1) -.17±.00 \\
 & LIME & \textbf{(5) .44±.00} & \textbf{(5) -.37±.00} & (0) NA & (0) NA & (0) NA & (0) NA & \textbf{(1) -.50±.00} \\
 & SHAP & \textbf{(5) -.29±.00} & (5) -.05±.00 & (0) NA & (0) NA & \textbf{(5) .54±.00} & \textbf{(5) -.42±.00} & \textbf{(3) -.38±.03} \\
\cmidrule{1-9}
\multirow[t]{6}{*}{GPT2} & Grad & \textbf{(3) .28±.00} & (3) .11±.00 & (2) -.12±.00 & (2) .19±.00 & \textbf{(5) .59±.56} & (0) NA & \textbf{(3) -.22±.09} \\
 & GxI & \textbf{(2) .66±.00} & (5) -.06±.30 & \textbf{(2) -.32±.00} & \textbf{(2) .42±.00} & \textbf{(5) .42±.50} & (0) NA & (4) -.18±.22 \\
 & IG & (0) NA & \textbf{(3) .36±.00} & (0) NA & \textbf{(2) .23±.00} & \textbf{(3) -.28±.00} & (0) NA & (2) -.18±.04 \\
 & IGxI & (3) .18±.00 & \textbf{(5) .54±.17} & (0) NA & \textbf{(2) .27±.00} & \textbf{(2) -.86±.00} & (5) -.01±.29 & (1) .11±.00 \\
 & LIME & \textbf{(3) -.31±.00} & \textbf{(3) .21±.00} & \textbf{(2) -.21±.00} & (2) .15±.00 & \textbf{(3) -.52±.00} & (3) .15±.00 & \textbf{(3) -.40±.09} \\
 & SHAP & \textbf{(5) -.43±.49} & \textbf{(5) .78±.06} & (0) NA & (0) NA & \textbf{(5) -.33±.35} & \textbf{(3) .28±.00} & \textbf{(2) -.52±.27} \\
\cmidrule{1-9}
\multirow[t]{6}{*}{TinyBERT} & Grad & \textbf{(5) .22±.00} & \textbf{(5) .90±.00} & (0) NA & (5) .15±.00 & \textbf{(5) .74±.00} & (0) NA & \textbf{(5) -.78±.21} \\
 & GxI & (0) NA & \textbf{(5) .85±.00} & (0) NA & (0) NA & \textbf{(5) .28±.00} & (0) NA & \textbf{(5) -.72±.19} \\
 & IG & (0) NA & \textbf{(5) .30±.00} & (0) NA & (0) NA & \textbf{(5) .46±.00} & \textbf{(5) -.25±.00} & \textbf{(5) -.69±.19} \\
 & IGxI & \textbf{(5) -.22±.00} & \textbf{(5) .77±.00} & (0) NA & (0) NA & \textbf{(5) -1.38±.00} & \textbf{(5) .64±.00} & \textbf{(5) -.78±.22} \\
 & LIME & (5) -.01±.00 & \textbf{(5) .52±.00} & (0) NA & (0) NA & \textbf{(5) -1.17±.00} & \textbf{(5) .40±.00} & \textbf{(5) -.62±.13} \\
 & SHAP & \textbf{(5) -.40±.00} & \textbf{(5) .66±.00} & (0) NA & (0) NA & (5) -.18±.00 & (0) NA & \textbf{(5) -.72±.16} \\
\cmidrule{1-9}
\multirow[t]{6}{*}{RoBERTa} & Grad & (5) .17±.37 & \textbf{(2) .28±.07} & (4) -.02±.02 & (4) .02±.02 & (4) -.02±.28 & (0) NA & (0) NA \\
 & GxI & (1) .05±.00 & \textbf{(2) .35±.03} & (1) .01±.00 & (1) -.02±.00 & (3) .11±.23 & (1) .10±.00 & (0) NA \\
 & IG & \textbf{(2) .36±.13} & (4) .07±.21 & (1) -.01±.00 & (3) .04±.04 & \textbf{(1) .23±.00} & (2) -.06±.17 & (0) NA \\
 & IGxI & \textbf{(3) .35±.17} & (4) .15±.29 & (1) .03±.00 & (1) -.05±.00 & \textbf{(2) .22±.52} & \textbf{(1) -.53±.00} & (0) NA \\
 & LIME & (5) .17±.15 & (4) .13±.52 & (4) -.03±.04 & (4) .03±.06 & (4) -.16±.28 & (4) .06±.25 & (3) .18±.02 \\
 & SHAP & (4) .04±.28 & (5) .11±.49 & (4) .00±.01 & (1) .01±.00 & (5) .15±.41 & \textbf{(4) -.23±.49} & \textbf{(1) .21±.00} \\
\bottomrule
\end{tabular}
\end{table}
\end{landscape}

\end{document}